\RequirePackage{fix-cm}
\documentclass[smallextended]{svjour3}       
\smartqed  
\usepackage{graphicx,amsmath,amssymb}
\usepackage{natbib}

\begin{document}

\title{An Overview of Distance and Similarity Functions for Structured Data
}


\author{Santiago Onta\~{n}\'{o}n}


\institute{Santiago Onta\~{n}\'{o}n \at
              	Google Research, Mountain View, CA 94043, USA \\
              \email{santiontanon@google.com}, and\\
	      	Drexel University, Philadelphia, PA 19104, USA\\
              \email{so367@drexel.edu}\\
}


\maketitle

\begin{abstract}
The notions of distance and similarity play a key role in many machine learning approaches, and artificial intelligence (AI) in general, since they can serve as an organizing principle by which individuals classify objects, form concepts and make generalizations. While  distance functions for propositional representations have been thoroughly studied, work on distance functions for structured representations, such as graphs, frames or logical clauses, has been carried out in different communities and is much less understood. Specifically, a significant amount of work that requires the use of a distance or similarity function for structured representations of data usually employs ad-hoc functions for specific applications. Therefore, the goal of this paper is to provide an overview of this work to identify connections between the work carried out in different areas and point out directions for future work.
\end{abstract}

\section{Introduction}\label{sec:intro}


The complementary notions of distance and similarity play a key role in many machine learning approaches, such as instance-based learning~\citep{aha1991instance}, kernel-based methods~\citep{vert2004primer}, case-based reasoning~\citep{Aamodt94}, or clustering algorithms~\citep{ng2002spectral,kaufman1987clustering}. Distance and similarity functions are also relevant for artificial intelligence (AI) in general, since they can serve as an organizing principle by which individuals classify objects, form concepts and make generalizations~\citep{tversky77similarity}. 
%
Specifically this paper presents an overview of distance and similarity functions for structured representations of data, such as graphs or frames. While  distance functions for propositional (i.e. feature-vector) representations have been thoroughly studied in the past, work on distance functions for structured representations has been carried out in different communities such as graph matching, inductive logic programming, case-based reasoning, relational learning or graph mining and is much less understood. Specifically, a significant amount of work that requires the use of a distance or similarity function for structured representations of data usually employs ad-hoc functions. Therefore, the goal of this paper is to provide an overview of this work in order to have a complete view of the field of distance functions for structure representations, and lay foundations for future work.

Structured data representations are important, since, there are many real-world application domains for which data of interest is inherently structured and it is hard to represent it using a propositional representation. Consider, for example, a biomedical domain where we are interested on predicting certain properties of chemical molecules. Representing molecules as feature vectors is problematic, since molecules can be of arbitrary sizes, but features vectors are fixed size. In this particular case, a graph-based representation might be able to more accurately represent the data of interest. 
Moreover, this paper only focuses on distance and similarity in the context of AI and machine learning. Psychological foundations of subjective assessments of similarity are out of scope. Interested readers are referred to the relevant cognitive science literature~\citep{tversky77similarity,holyoak1987surface,goldstone1991relational}. Additionally, while methods for similarity and distance assessment are related to areas such as {\em ontology alignment}~\citep{kalfoglou2003ontology} or {\em computational analogy}~\citep{french2002computational}, here we will focus on the core techniques, and will not discuss applications to ontology alignment, or other areas.

The remainder of this paper is structured as follows. Section \ref{sec:background} provides some necessary background. After that, the paper overviews the existing literature by dividing the body of work into three large classes of structured representations: distance functions for graph-based representations are discussed in Section \ref{sec:graph-distance}, those for logic-based representations are discussed in Section \ref{sec:logic-distance}, and finally Section \ref{sec:frame-distance} focuses on functions for frame-based representations. Section \ref{sec:discussion} discusses connections between those areas of work, and the paper closes with conclusions and future research directions.


\section{Background}\label{sec:background}

This section presents some basic concepts of distance and similarity functions, as well as of structured data representations.

\subsection{Distance and Similarity Functions}\label{sec:similarity-bg}

Many machine learning and AI methods require assessing how similar or how different two objects are. For example, the $k$-nearest neighbor algorithm~\citep{cover1967nearest} uses a distance function to determine, out of all the instances in the training set, which ones are the most similar to the target, to then predict a label for it, given the labels of the $k$ most similar instances. Intuitively, {\em distance functions} are mathematical functions that assign a numerical value (their {\em distance}) to each pair of objects in a given domain. This numerical value represents an assessment of how similar they are: two very similar objects would be assigned a very low distance, and two very dissimilar objects would be assigned a larger distance. Similarity functions are the complementary idea, and assign high similarity values to similar objects, and low values to dissimilar pairs of objects. 

\begin{definition}[distance metric]\label{def:distance}
A {\em distance metric} $d$ over objects in a set $X$ is a function: $d : X \times X \to [0,\infty)$ such that, for each $x, y, z \in X$ the following properties are satisfied:
\begin{itemize}
\item $d(x,y) \geq 0$ (Non-negativity)
\item $d(x,y) = 0 \iff x = y$ (Identity)
\item $d(x,y) = d(y,x)$ (Symmetry)
\item $d(x,z) \leq d(x,y) + d(y,z)$ (Triangle inequality)
\end{itemize}
\end{definition}

Some definitions replace {\em non-negativity} by a {\em minimality} property: $d(x,y) \geq d(x,x)$. Since $d(x,x) = 0$ due to the {\em identity} property, these are equivalent. Moreover, although in mathematics, the terms {\em distance}, {\em metric} and {\em distance function} are synonyms, in this paper, we will use following convention: 
\begin{itemize}
\item we will use the term {\em distance metric} to refer to a function that satisfies the above definition, 
\item we will use the term {\em distance measure} to refer to a function that intuitively captures the notion of ``distance between objects'', but does not satisfy at least one of the four properties in Definition \ref{def:distance},
\item finally, we will use the term {\em distance function} as the general term to denote either distance metrics or distance measures. We will also use the term {\em distance functions} to refer to {\em similarity and distance functions} when context allows, in order to avoid repeating ``similarity and distance functions'' constantly.
\end{itemize}

Often, some of the properties above are not required (for example, most algorithms would not be affected if the distance function used does not satisfy the triangle inequality). However, when defining distance functions, it is important that they satisfy all four properties, since some algorithms (e.g., the classic Fish'n'Shrink~\citep{schaaf1996fish}) assume that the distance function used is a metric. Moreover, some authors have argued that in some application domains, where we want the distance or similarity function to approximate perceptual similarity as would be judged by a human, these mathematical properties provide too rigid a framework, and other, alternative properties ({\em dominance}, {\em consistency} and {\em transitivity}) have been proposed~\citep{santini1999similarity}. 

Although there is no agreed upon definition of similarity function in the literature, in the rest of this paper, we will use the following definition. 
\begin{definition}[similarity function]\label{def:similarity}
A {\em similarity function} $s$ over objects in a set $X$ is a function: $s : X \times X \to [0,u]$, where $u$ is an upper bound (i.e., the maximum similarity value, usually $u = 1$), and where for each $x, y \in X$ the following properties are satisfied:
\begin{itemize}
\item $d(x,y) \geq 0$ (Non-negativity)
\item $d(x,y) \leq u$ (Boundedness)
\item $s(x,y) = u \iff x = y$ (Identity)
\item $s(x,y) = s(y,x)$ (Symmetry)
\end{itemize}
\end{definition}

Intuitively, a similarity function is the complementary concept to a distance function. For each distance function $d$, we can define its associated similarity function as $s_d(x,y) = u/(1+d(x,y))$. Other than being complementary functions (when distance grows, similarity decreases), usually distance functions are unbounded, whereas similarity functions are bounded to a range $[0,u]$, and also, there is no equivalent property to the triangle inequality for similarity functions, and thus, they are not {\em metrics} in the mathematical sense. Moreover, similarly as for distance functions, when similarity functions are used to capture perceptual similarity, some authors have argued that the {\em symmetry} property should be dropped, as human perception of similarity seems not to be symmetric~\citep{tversky77similarity}.

\subsection{Standard Methods to Assess Distance and Similarity}\label{sec:basic-similarities}

Because of their importance in AI and other fields, a very large number of distance and similarity functions have been defined in the literature. Since many distance functions for structured representations are based on more basic notions of similarity between basic representations such as vectors or strings, this section presents a list of the most common ways to assess similarity between non-structured data representations (a summary of the most common functions can be seen in Table \ref{tbl:basefunctions}).

\begin{table}[tb]
    \centering
    \small{
    \begin{tabular}{|l|l|} \hline
    	Data Representation	&	Common Distance and Similarity Functions \\ \hline \hline
	Scalars/Vectors		&	Minkowski (Manhattan, Euclidean, Chebyshev)	\\
					&	Cosine Similarity	\\ \hline
	Sets				&	Tverski \\
					&	Jaccard Index \\
					&	S{\o}rensen's Index (Dice coefficient) \\ \hline
	Sequences		&	Edit Distances (Levenshtein)	\\
					&	Sequence Alignment	\\
					&	Dynamic Time Warping	\\
					&	Auto-Regressive Measures		\\
					& 	Compression Distance	\\ \hline
	Hierarchies/Taxonomies	&	Rada (edge counting) \\
						&	Resnik (information content) \\ \hline
	Probability Distributions	&	KL Divergence	\\
						&	Wasserstein Metric \\ \hline
    \end{tabular}
    }
    \caption{Common distance and similarity functions for non-structured data representations.}
    \label{tbl:basefunctions}
\end{table}

\subsubsection{Scalars and Vectors}

The most common distance functions between scalars and vectors are the different instantiations of the {\bf Minkowski distance}, and the {\bf cosine similarity}:

\[d_{\mathit{Minkowski}}(\vec{x},\vec{y}) = \left(\sum_{i=1...n} |x_i - y_i|^p\right)^{\frac{1}{p}} \]

When $p = 1$ we have the Manhattan distance, when $p = 2$ we have the Euclidean distance, and when $p = \infty$ it converges to the Chebyshev distance ($d_{\mathit{Chebyshev}}(\vec{x},\vec{y}) = max_{i = 1 ...n} |x_i - y_i|$). Also, when $n=1$ (i.e., when comparing scalars), this corresponds to the absolute value of their difference.

The {\em cosine similarity}~\citep{singhal2001modern} measures the cosine of the angle between two vectors and is defined as:

\[s_{\mathit{cosine}}(\vec{x},\vec{y}) = \frac{\vec{x} \vec{y}}{|\vec{x}| |\vec{y}|}\]

Intuitively, the cosine similarity differs from the Minkowski distance in that the magnitude of the vectors is not considered, and only the angle between them is measured (if they both point in the same direction, the cosine of the angle is 1, and if they are orthogonal, the cosine of their angle is 0). This gives them different semantics, making them appropriate in different applications.

\subsubsection{Sets}\label{subsec:sets}

The most well known measures are {\bf Tverski's}~\citep{tversky77similarity}, the {\bf Jaccard index}, or {\bf S{\o}rensen's index}~\citep{sorenson1948method} (also known as Dice's coefficient), with Jaccard being the most common. Given two sets $X$ and $Y$, Tverski's index is defined as:

\[s_{\mathit{Tverski}}(X,Y) = \frac{|X \cap Y|}{|X \cup Y| + \alpha|X-Y| + \beta|Y-X|}\]

Whereas Jaccard's index is the special case where $\alpha = \beta = 0$:

\[ s_{\mathit{Jaccard}}(X,Y)=  \frac{|X \cap Y|}{|X \cup Y|}\] 

Intuitively, this results in a similarity of 1 if both sets are identical (since the size of their intersection and union would be the same), and a similarity of 0 for two disjoint sets. Variations of these measures exist, such as the {\em continuous Jaccard index} where elements could belong to a set with a certain degree represented by a real number~\citep{valls2014toward}.

\subsubsection{Sequences}\label{subsec:sequences}

Many distance functions exist for comparing sequences. The most common family of distances is that of {\bf edit distances}, where the distance between two sequences is defined as the number of ``edit operations'' that one needs to perform to one sequences in order to obtain the second. The most common edit distance is the {\bf Levenshtein distance}~\citep{levenshtein1966binary}, where there are only three edit operations allowed: {\em insertion} (inserting a symbol into the sequence), {\em deletion} (removing a symbol from the sequence) and {\em replacement} (replacing a symbol by another symbol). For example, if we consider words as sequences of letters (i.e., strings) the distance between ``hello'' and ``mellow'' is 2, wince we can {\em replace} the ``h'' by an ``m'' and then {\em insert} an ``w'' at the end. Extensions exist where different edit operations have different weights, or where additional edit operations (such as transpositions) are allowed. Distances such as the {\bf longest common subsequence} can also be seen as edit distances (with just {\em insertion} and {\em deletion} as the edit operations).

Another very common approach is that of {\em sequence alignment}~\citep{gollery2005bioinformatics}, which is very common in biological domains due to the obvious application of comparing DNA sequences. Specifically, the problem of calculating a {\em global alignment} between two sequences is equivalent to the problem of calculating the edit distance, and thus both approaches share algorithms, with the {\bf Needleman-Wunsch algorithm}~\citep{needleman1970general} being the most common. The only difference between edit distance and alignment is that when we want to output an alignment, the algorithm needs to keep a ``back trace'' so that we can then output which elements fro one sequence correspond to which other elements of another sequence. A very common alignment algorithm used in time series matching is {\bf Dynamic Time Warping}~\citep{itakura1975minimum}, which uses a dynamic programming approach with very small differences with respect to Needleman-Wunsch's algorithm.

{\bf Auto-regressive measures} are based on learning probabilistic models of sequences, and then comparing the sequences by comparing the parameters of the learnt models. For example, \cite{ramoni2002bayesian} propose an approach to cluster time series based on training a Markov chain for each sequence, and then using the KL divergence~\citep{kullbackLeibler1951divergence} as a similarity function between the trained Markov chains as a similarity function between time series. This idea has also been used to compare agent behaviors in the context of learning from demonstration~\citep{ontanon2014dynamic}.

Finally, another common idea is that of information content. The underlying idea of these approaches is the notion of {\em Kolmogorov complexity}~\citep{kolmogorov1965three}: the Kolmogorov complexity of a string is the length in bits of the smallest program that can generate such string as output (e.g., the length of the description of the smallest Turing machine that generates such string). One idea is to compute the Kolmogorov complexity of computing one string when the other is given as an auxiliary input (notice that this is also related to the idea of edit distance). Given that the Kolmogorov complexity is not computable, a common approximation is to use a compression algorithm $C$ (such as LZW~\citep{welch1984technique}) as an approximation. This leads to the {\bf normalized compression distance}~\citep{cilibrasi2005clustering}:

\[d_{NCD}(x, y) = \frac{C(xy) - min\{C(x), C(y)\}}{max\{C(x), C(y)\}}\]


\noindent where $C(x)$ is the size of the resulting compressed version of the sequence $x$, and $C(xy)$ is the size of the compressed version of concatenating $x$ and $y$. Intuitively, the compression algorithm is used for two purposes: $C(x)$ approximates the Kolmogorov complexity of a sequence, and $C(xy) - min\{C(x), C(y)\}$ approximates the length of the smallest program to generate one sequence given the other as an auxiliary input. Also, if $x$ and $y$ are very similar, then compressing $xy$ should have almost the same size than compressing one of them. 

For the particular case of numerical sequences (time series), a number of specialized distance functions have been developed beyond those described above. For example, re-sampling the two time series and using a distance function between vectors (e.g., Euclidean)~\citep{keogh2003need}, using Fourier transform coefficients~\citep{agrawal1993efficient}, {\em time-warped edit distance}~\citep{marteau2009time}. For a comparison between these measures, the reader is referred to the work of~\cite{serra2014empirical}.

\subsubsection{Hierarchies or Taxonomies}\label{subsec:hierarchies}

Distance functions between elements in a hierarchy are also a common source for defining distance functions for structured representations. A hierarchy is defined as a partially ordered set  $\langle X, \leq \rangle$ with elements $X$ ordered by a relation $\leq$, where each element in $X$ has at most one parent. We say that $x$ is the {\em parent} of $y$ if $x \leq y$, and $\nexists x' \in X : x' \neq X \,\,\, \wedge \,\,\, x \leq x' \leq y$. Usually hierarchies have a special element $x_\bot \in X$ such that $\forall x \in X: x_\bot \leq x$. $x_\bot$ is called the {\em root} of the hierarchy, or the {\em bottom} element. Common examples of hierarchies are class hierarchies in object oriented programming, or some of the different classifications of words in Wordnet~\citep{miller1995wordnet} such as {\em hypernyms}.

The most common distance functions between elements in a hierarchy are:
\begin{itemize}
\item {\bf Rada's}~\citep{rada1989development} (often referred to as ``edge counting''): in this distance function a hierarchy is seen as a tree, where the {\em parent} relation defines the edges between the elements in the tree: 
\[d_{rada}(x,y) = |path(x,z)| + |path(y,z)|\]
\noindent where $z$ is the deepest element in the hierarchy of which both $x$ and $y$ are descendants, and $path(x,z)$ is the number of edges that need to be traversed to reach $z$ from $x$. $z$ is also known as the {\em least general generalization}, when $\leq$ is considered to be a {\em more general than} relation.
\item {\bf Resnik's}~\citep{resnik1995using,resnik1999semantic} (often referred to as ``information content''): in this similarity function, elements in the hierarchy are seen as {\em concepts}, and we have access to a function $p : X \to [0,1]$, which, given a concept in the hierarchy, gives us the probability of encountering an {\em instance} of such concept:
\[s_{resnik}(x,y) = max_{z \in X | z \leq x \,\, \wedge \,\, z \leq y} [-log p(z)]\]
\noindent
the main difference between Rada's and Resnik's distance functions is that Rada's measure considers each edge in the hierarchy to count the same toward the distance of two concepts, while Resnik's takes into account that some edges are more important than others. For example, if there are two concepts $x$ and $y$,w where $x$ is the parent of $y$, but where $y$ is almost identical to $x$, then $p(x)$ will also be very similar to $p(y)$, and thus, the edge between them will have little weight in similarity calculations. In practical applications $p$ can be estimated from a training set of instances.
\end{itemize}

Other measures in the literature integrate ideas from these two basic functions. For example, \cite{jiang1997semantic} define a distance function that integrates both edge counting and information content, showing good results. Wu and Palmer's conceptual similarity function for concepts in WordNet is basically a normalized version of Rada's function~\citep{wu1994verbs}.

\subsubsection{Probability Distributions}

The most common way to compare probability distributions is probably using the {\bf Kullback-Leibler Divergence} (KL divergence)~\citep{kullbackLeibler1951divergence}, defined for two probability distributions $Q$ and $P$ as follows: 

\[d_{KL}(P,Q) = - \sum_i P(i) log \frac{Q(i)}{P(i)}\]

Intuitively this measures the amount of information lost when one uses $Q$ to approximate $P$. When considering continuous distributions, we just need to replace the discrete sum by an integral operation. Also, notice that the Kullback-Leibler Divergence is not a distance metric, since it does not satisfy the {\em symmetry} or the {\em triangle inequality} properties. 

Another very common distance metric is the {\bf Wasserstein metric}~\citep{dobrushin1970prescribing} (also known as the {\em earth-mover distance}~\citep{rubner2000earth}). The intuitive idea is to see two probability distributions as two different ways to pile dirt over an {\em area}, and then calculate the amount of work required to move dirt to turn one distribution into the other. Thus, this metric requires as input parameter a distance function that defines the {\em area} (or a pair-wise distance matrix in the case of discrete distributions), that indicates the amount of work of moving dirt between two points in this area. Calculating this distance requires solving a linear optimization problem to find the optimal ``flow'' of dirt. This distance has also been used to compare images in the context of image retrieval~\citep{rubner2000earth};

Other common measures include the $\chi^2$ statistic, among others. As mentioned in Section \ref{subsec:sequences}, these distance functions have been also been used to calculate distance or similarity between sequences by first representing the sequences as {\em stochastic processes}, and then comparing the probability distributions that govern these processes. 

\subsubsection{Weighting and Metric Learning}\label{subsec:metriclearning}

Notice that all the distance functions presented above just calculate a numerical distance or similarity between two objects without having in mind the problem at hand. When using these functions as part of a machine learning algorithm, e.g., $k$-nn, it might be desirable to use the information from the labeled data in the training set to adjust the distance function to the problem at hand. For example, some variable might be completely irrelevant for the prediction task at hand, and we would not want that variable to play any role in the distance calculations. Many distance functions allow for such process. For example, the {\bf Mahalanobis distance} generalizes the Euclidean distance by calculating a covariance matrix from the training data. Edit distances can be extended by defining different weights for the different edit operations. In general this idea is studied in several subfields of machine learning such as {\em metric learning}~\citep{kulis2013metric,bellet2013survey} or {\em feature weighting and selection}~\citep{wettschereck1997review}. Some distances, such as the normalized compression distance, do not easily allow for metric learning.

\subsection{Structured Data Representations}\label{sec:structured-bg}

The vast majority of machine learning approaches in the literature uses feature-vector representations of data (i.e., {\em propositional representations}) where each training instance is represented as a fixed-size vector/tensor of either numeric or categorical values. 
Moreover, a very large number of structured representations has been proposed in the literature, which will be the focus of this paper. We will group them into three major categories: graph-based representations, logic-based representations, and frame-based representations. We briefly describe these representations below.

\subsubsection{Graph-based Representations}

Graph-based representations use graphs in various ways to represent instances. A common approach is to use {\bf directed labeled graphs} (DLGs).

\begin{definition}[Directed Labeled Graph]
Given a finite set of labels $L$, a directed labeled graph $g$ is defined as a tuple $g = \langle V, E, l \rangle$, where:
\begin{itemize}
\item $V = \{v_1, ..., v_n\}$ is a finite set of vertices,
\item $E = \{(v_{i_1}, v_{j_1}), ..., (v_{i_m}, v_{j_m})\}$ is a finite set of directed edges,
\item $l : V \cup E \to L$ assigns a label from $L$ to each vertex or edge.
\end{itemize}
\end{definition}

\begin{figure}[tb]
    \centering
    \includegraphics[width=\textwidth]{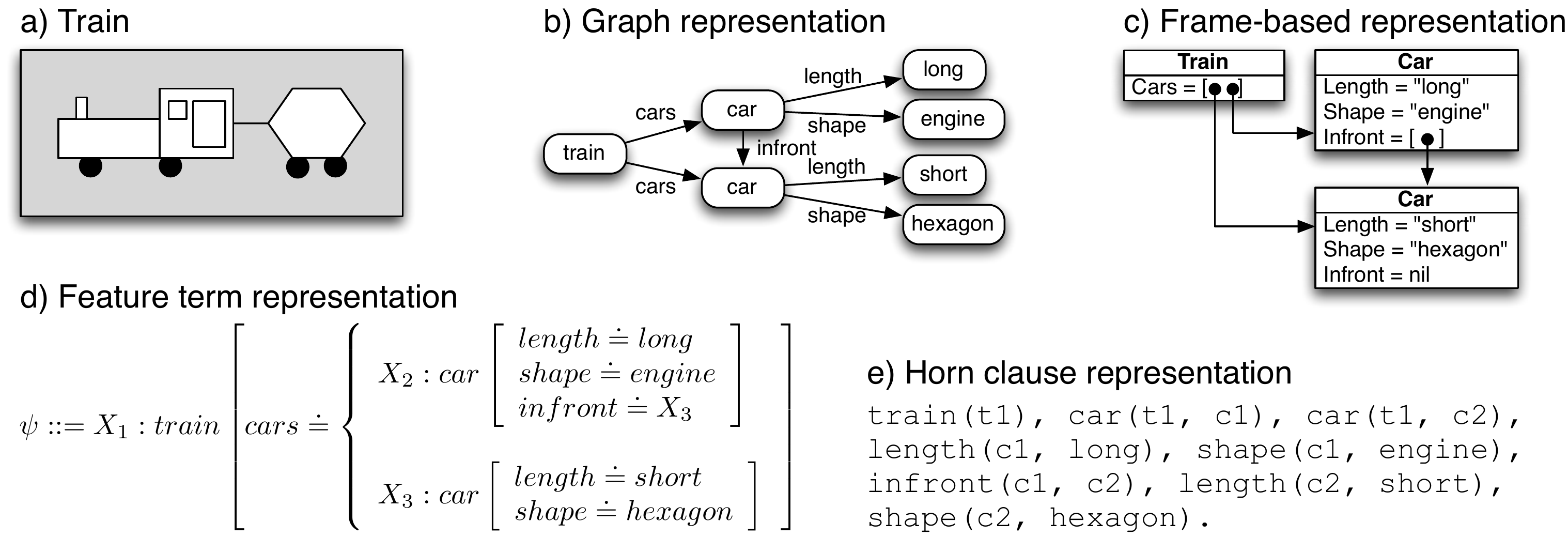}
    \caption{A small train (inspired by Michalski's train dataset~\citep{michalski77trains}, represented in different structured representation formalisms.}
    \label{fig:representations}
\end{figure}

For example, a lot of work in machine learning applied to biochemical domains use labeled graphs to represent molecules~\citep{bunke1998graph,Kashima2003kernel,gartner2003survey,mahe2005graph}. Another common example of graph-based data is in computer vision, where graphs and graph matching algorithms have been extensively used for many tasks such as character or 3d object recognition~\citep{bunke2000graph}. We can distinguish two different ways to use graphs in structured machine learning: 

\begin{itemize}
\item The {\em one graph-per instance} approach: where each training instance is represented by a complete graph (e.g. a chemical molecule, where each vertex is an atom, and edges represent chemical bonds). Figure \ref{fig:representations}.b shows an example of this approach, representing a small train inspired in the classic trains dataset by~\cite{michalski77trains}, using a DLG. 
\item The {\em one vertex per instance} approach: where each vertex in the graph represents an instance, and edges represent their relationships. For example,  users in a social network (where each vertex is a user, and edges represent connections or other relations), scientific paper citation graphs or web pages are commonly represented this way.
\end{itemize}

The main difference from the point of view of distance functions is that the former requires distance functions between graphs, and the later distance functions between vertices on a graph. In many approaches the basic theoretical graph definition is extended or slightly modified. For example, {\em conceptual graphs}~\citep{sowa1979semantics} are hierarchical bipartite graphs where some vertices represent entities and some other represent relations, and where a vertex could contain a nested subgraph inside of it. Another example are {\em attributed graphs}, where each vertex can have a collection of numerical or symbolic features~\citep{tsai1979error}. For example, when representing molecules as attributed graphs, we could have features representing the distances of each bond or the coordinates of each atom in a 3D space~\citep{riesen2008iam}.

\subsubsection{Logic-based Representations}

Logic-based representations have been studied for decades in the {\em inductive logic programming}~\citep{lavrac1994inductive} community, as well as in {\em explanation-based learning}~\citep{mitchell1986explanation}. The key idea is to represent the training data using logical clauses. For example, Figure \ref{fig:representations}.d and Figure \ref{fig:representations}.e shows how a small object represented using two different logical formalisms (feature terms and Horn clauses). Most of these logic-based formalisms correspond to different subsets of {\em first order logic} (FOL). The most common are:

\begin{itemize}
\item {\bf Horn clauses}: these are the most common representation in the ILP community. Objects are represented as conjunctions grounded terms (such as the example in Figure Figure \ref{fig:representations}.e). Some times, the object representation is augmented with a theory expressed as Horn clauses, which can be used to draw additional inferences on the objects being represented. 
\item {\bf Description Logics}~\citep{DL2003handbook}: common in the {\em semantic web} community~\citep{baader2005description}, description logics (DLs) are a family of knowledge representation languages that correspond to different subsets of FOL. Many different DLs exist, representing different tradeoffs between representation power, and computational tractability. The original purpose of DLs was to provide formal semantics to frame-based representations and semantic networks.
\item {\bf Feature Terms} (also known as {\em order-sorted feature structures}, or feature logics)~\citep{Carpenter1992}: another subset of FOL, that has been frequently used in the case-based reasoning community~\citep{plaza1995cases} and in natural language processing~\citep{emele1990typed,krieger2011efficient,shieber2003introduction}. An example train represented as a feature term is shown in Figure \ref{fig:representations}.d.
\end{itemize}

The main strength of logic-based representations is that the can naturally encode the concept of generalization (via subsumption operations), and inference, and that they naturally allow for background knowledge to be represented in the form of rules. For example, many learning systems based on logic-based representations utilize the concept of {\em least-general generalization}~\citep{Plotkin70} to induce rules, or even assess distance and similarity between instances~\citep{ontanon2012similarity,SanchezRuizGranadosOGP11}.

\subsubsection{Frame-based Representations}\label{subsec:frames}

We will group all the different representations that derive from the original idea of {\em frames}~\citep{minsky1974framework} as frame-based representations. These representations are common in fields such as {\em case-based reasoning} (CBR)~\citep{Aamodt94} and {\em statistical relational learning} (SRL)~\citep{getoor2007introduction}. While in some CBR work, the distinction is made between {\em frame-based representations} and {\em object-oriented representations}, and the former is associated with description logics~\citep{bergmann2005representation}, given the nature of the existing work on distance and similarity assessment, we will use the term {\em frame-based representation} to capture all representations such as object-oriented ones, whose main constructs are ``is-a'' and ``part-of'' relations. An example such representation is shown in Figure \ref{fig:representations}.c.

Frame-based representations are very common in the CBR literature, specially on the early days of CBR with systems like CASUEL~\citep{manago1994casuel} or HOMER~\citep{goker1999development}. Often, these representations are seen as direct generalizations of flat feature-vector representations, where objects are represented by a set of attribute-value pairs (called {\em slots}), determined by an object {\em class} (where classes form a hierarchy). The attributes of these objects can be ``simple attributes'' (which take numerical or symbolic values) or ``relational attributes'' (which take other objects for values)~\citep{bergmann1998similarity}. 
This idea of simple and relational attributes is analogous to the idea of attributes and relationships in {\em entity-relationship} (ER)~\citep{chen1988entity} models, commonly used to define relational databases. In the field of SRL, extensions of the ER model to account for probabilistic data (probabilistic entity-relationship models, or PER) are common, with a well known example being DAPER~\citep{heckerman2007probabilistic}.


\section{Distance Functions for Graph-based Representations}\label{sec:graph-distance}

This section provides an overview of the large amount of work  existing in the literature on distance and similarity functions for graph-based representations. We classify this work in four main types of functions: those based on {\em graph matching}, those based on the idea of {\em edit distance}, those based on {\em refinement graphs}, and finally the attempts to encapsulate these functions as {\em kernels}. Additionally, we will see how some of these ideas are related (for example, we will see that certain types of graph matching operations are actually equivalent to edit distances under certain cost functions).

\subsection{Graph Matching-based Distance Functions}

Work on graph matching can be traced back to the 1960s with pioneering work on graph isomorphism by \cite{sussenguth1964structure}. Significant contributions to the field have been done since. This section will cover only the work concerning distance and similarity assessment, for comprehensive overview of the field, the reader is referred to recent overviews by~\cite{conte2004thirty} or \cite{emmert2016fifty}. 

The most stringent formulation of the graph matching is the well known {\bf graph isomorphism} problem~\citep{babai2018groups}. Given two graphs $g_1$ and $g_2$, the problem is to find a bijection $f$ (i.e., a one-to-one correspondence) between the vertices of both graphs such that two vertices $v$ and $v'$ are adjacent in $g_1$, if and only if $f(v)$ and $f(v')$ are adjacent in $g_2$. Moreover, if graphs are labeled, then the labels of $v$ and $f(v)$ must also match, as well as the labels of the edges between $v$ and $v'$ and between $f(v)$ and $f(v')$. Intuitively, this amounts to checking if two graphs are identical structurally. As of the writing of this document, the complexity of graph isomorphism has not yet been determined, but it has been recently conjectured to be quasipolynomial by \cite{babai2018groups}. While graph isomorphism is not particularly useful for the purpose of distance calculations, relaxations of this problem have been used extensively for assessing distance and similarities between graphs.

The immediate relaxation of graph isomorphism is what is known as {\bf subgraph isomorphism}~\citep{read1977graph}, corresponding to finding if there is a graph isomorphism between a graph $g_1$ and any subgraph of another graph $g_2$. A further relaxation is the {\bf maximally common subgraph} (MCS)~\citep{levi1973note}, which is particularly interesting for distance and similarity assessment.  The MCS problem consists on finding what is the largest subgraph of $g_1$ for which we can find a subgraph isomorphism with respect to $g_2$. Distance functions for graphs based on the MCS include (all these three functions are distance metrics):
\begin{itemize}
\item \cite{bunke1998graph} showed that the following distance function based on the size (in vertices) of the MCS is a metric:
\[d_{bs}(g_1, g_2) = 1 - \frac{|MCS(g_1, g_2)|}{max(|g_1|,|g_2|)}\]
\item \cite{wallis2001graph} proposed a variation over Bunke and Shearer's distance normalizing by the size of the union graph, rather than by the size of the larger graph:
\[d_{wskr}(g_1, g_2) = 1 - \frac{|MCS(g_1, g_2)|}{|g_1 \cup g_2|}\]
\noindent where $|g_1 \cup g_2|$ is calculated as $|g_1| + |g_2| - |MCS(g_1, g_2)|$. Thus, notice that if we interpret the MCS as the {\em union} of two graphs, this distance is basically the Jaccard distance (see Section \ref{subsec:sets}), applied to graphs.
\item \cite{fernandez2001graph} propose a different variant that involves calculating both the MCS and the {\em mcs} (minimum common supergraph) (which we will write in lower case, to distinguish from the MCS, and corresponds to the minimum graph $g$ such that we can find a subgraph isomorphism between both $g_1$ and $g_1$ and $g$):
\[d_{fv}(g_1, g_2) = |mcs(g_1, g_2)| - |MCS(g_1, g_2)|\]
\end{itemize}

However, both subgraph isomorphism and the MCS problem are known to be NP-complete~\cite{bunke1997relation}. The original algorithm by \cite{levi1973note} had a complexity of $O((nm)^n)$ (where $n$ and $m$ are the number of vertices of the two graphs), and the more recent algorithm by \cite{abu2007maximum} is $O(3^{m/3}(m+1)^c)$, where $c$ is the size of the smaller vertex cover between the two inputs. Therefore, methods based on approximations of the MCS have algo been proposed. For example, {\bf MatchBox}~\citep{schadler1999comparing} uses Hopfield-style neural networks to approximate MCS-based graph matching distances between two labeled graphs. 

There is also a significant amount of work on defining distance functions between graphs using graph matching techniques using slightly different criteria than strict (sub)graph isomorphism or MCS calculations. Graph isomorphism requires finding a mapping between two graphs that satisfies a specific set of criteria. If we relax or modify these criteria, a range of different distance functions can be defined. For example:
\begin{itemize}
\item Some early work on graph matching by \cite{shapiro1981structural} proposed the idea of finding {\bf $\epsilon$-homomorphisms} between hypergraphs (they considered graphs with vertices and ``relations'', where ``relations'' could involve 2 or more vertices). Where $\epsilon$ is a measure of dissimilarity beteween 0 and 1. Assuming the existence of a weighting function for each element in a graph (vertices and relations) such that all the weights add up to 1, there is an $\epsilon$-homomorphisms between two graphs if we can find a mapping such that the sum of the elements in the graphs that are mot matched is less than $\epsilon$. In order to solve this problem, they proposed to use systematic search using backtracking.

\item \cite{poole1995novel} propose a variation of the MCS approach, where they find the {\em most interesting common generalization} (MICG), defined as the generalization of two graphs that maximizes a user-provided measure of {\em interest} (which must satisfy certain properties, such as not to increase if edges or vertices are removed). The similarity between two graphs, is then defined as:
\[s_{pc}(g_1, g_2) = \frac{interest(MICG(g_1, g_2))}{max(interest(g_1), interest(g_2))}\]
In order to find the MICG of two graphs, they employ $A^*$ search over the product graph of $g_1$ and $g_2$ to find a consistent subgraph that maximizes the function of interest.

\item The similarity function proposed by \cite{champin2003measuring} for multi-labeled graphs (each vertex or edge can have one or more labels) differs from the standard MCS-based approaches above in two key ways: 1) they allow for a used-specified function $f$ to score the mapping (rather than finding the mapping that finds the MCS), and 2) they do not require the mapping from vertices of one graph to the other graph to be one-to-one. Their proposal similarity function is as follows:
\[s^{cs}_m(g_1, g_2) = \frac{f(g_1 \sqcap_m g_2) - g(splits(m))}{f(g_1 \cup g_2)}\]
\noindent where $splits$ measures the number of non one-to-one mappings (assuming that we want to penalize this), $g_1 \sqcap_m g_2$ is the intersection graph, given the mapping $m$ (i.e., a graph containing only those matched vertices and edges), and $f$ and $g$ are user-defined functions. In order to assess similarity, they propose a greedy algorithm to find the mapping $m$ that maximizes this similarity.

\item \cite{wang1997method} propose another similar measure, assuming the existence of a function $W$ that assigns an importance {\em score} to each vertex and edge. Given a mapping $m$, $W$ can be used to define the similarity of two graphs as follows. For each vertex $v$ in graph $g_1$ that is mapped to a vertex $m(v)$ in $g_2$, the score of this mapping is the average of $W(v)$ and $W(m(v))$ (score for edges is analogous). Let us call $F_v$ to the sum of the scores of all the vertices, and $F_e$ to the sum of the score of all the edges, and $M_v$ and $M_e$ the maximum score for vertices and edges that is theoretically possible given the number of vertices of $g_1$ and $g_1$ and the range of $W$. Similarity is then assessed as:
\[s^{wi}_m(g_1, g_2) = \frac{F_v + F_e}{M_v + M_e}\]
\item In order to find the mapping that maximizes this function, they propose the use of a genetic algorithm.
\end{itemize}

Other examples include the work of \cite{mishne2004source}, where they do not impose some of the usual isomorphism constraints on the mapping they find, and just mapping each vertex to the most similar vertex on the other graph, given a constrained neighborhood with radius $n$, making the problem $O(n^3)$. The use this approach to develop a similarity function to retrieve source code, representing it as conceptual graphs.

Finally, the concept of graph matching is related to the idea of {\em analogical mapping}. For example, in order to calculate analogical mappings, \cite{leishman1989analogy} compute what they call {\em minimal common generalization} of two graphs, which is a similar concept to the MCS, except that instead of calculating the maximum subgraph, they calculate the maximum subgraph that maximizes some measure of analogical mapping score. A very well known approach related to this is the {\bf structure mapping engine} (SME)~\citep{falkenhainer1989structure}, which calculates analogical mappings that maximize a scoring function based on structure mapping theory~\citep{gentner1983structure}. The concept of analogical mapping is very related to that of similarity~\citep{holyoak1987surface}, and specifically, the score used by SME has been used in the literature as a measure of similarity between graphs~\citep{ontanon2011sam}. 

To conclude this section, we would like to point out relations to other ideas of distance and similarity. Specifically:
\begin{itemize}
\item The idea of calculating the MCS, or some variant, and use a measure of size on it to assess similarity or distance between graphs is both related to the idea of the Jaccard similarity (as pointed out above), as well as to the idea of distance functions in hierarchies. If we see each graph as an element of a hierarchy, and the {\em subgraph-isomorphism} relation as the {\em parent} relation, then many of the ideas of similarity presented in this section can be seen as versions of {\em Rada's} or {\em Resnik's} distances presented in Section \ref{subsec:hierarchies} (with measures based on the size of the MCS being related to Rada's and measures, such as Poole and Campbells, based on information content or interest, related to Resnik's). This will be made more clear below in Section \ref{subsec:graph-refinement}.
\item It has been shown in the literature that the problem of calculating the MCS, is a special case of calculating the {\em edit distance} between graphs~\citep{bunke1997relation} (described below).
\end{itemize}

\subsection{Graph Edit Distance Functions}

The idea of adapting the edit distance (described in Section \ref{subsec:sequences}) to graphs) can be traced back to the early work of~\cite{sanfeliu1983distance}. The basic idea is the following. Given two graphs $g_1 = \langle V_1, E_1, l_1 \rangle$ and $g_2 = \langle V_2, E_2, l_2 \rangle$, let $m : V'_1 \to V'_2$ be a bijective mapping between a subset of vertices $V'_1 \subseteq V_1$ of $g_1$ and a subset of vertices $V'_2 \subseteq V_2$ of $g_2$. We will call $E'_1 \subseteq E_1$ to the subset of edges of $g_1$ involving vertices in $V'_1$, and define $m((v_1,v_2) = (m(v_1), m(v_2))$. The cost of a mapping $m$ is defined as:
\[\gamma(g_1, g_2, m) = \sum_{v \in V_1 - V'_1} c_d(v) + \sum_{w \in V_2 - V'_2} c_i(w) + \sum_{v \in V'_1} c_s(v,m(v)) + \sum_{e \in E'_1} c_s(e, m(e)) \]
\noindent where $c_d$, $c_i$, $c_s$, and $c_s$ are predefined cost functions for deleting vertices, inserting vertices, substituting a vertex by another, and substituting an edge by another, respectively. The cost of the optimal mapping $m$ (the one with the lowest cost) is called the {\bf graph edit distance} between $g_1$ and $g_2$~\citep{bunke1999error}. Calculating the edit distance is NP-complete~\citep{bunke1997relation}, and is usually done using tree search algorithms. Additionally, as \cite{bunke1999error} demonstrated, graph isomorphism, subgraph isomorphism and finding the MCS are special cases of calculating the edit distance under particular cost functions. 

Given the high computational complexity of the graph edit distance, several approaches exist to attempt to approximate it via different types of simplifications. For example, \cite{riesen2009approximate} propose an approximate graph edit distance approach based on the {\bf Hungarian algorithm}~\citep{munkres1957algorithms}, with polynomial complexity ($O(n^3)$, where $n$ is the number of vertices in the graphs). The Hungarian algorithm is designed to solve the {\em assignment problem}, i.e., given a set of $n$ ``variables'', each of which can take $m$ different ``values'', and where we have a {\em cost matrix} specifying the cost of assigning each different value to each different variable, finding the optimal value assignment to each variable, such that no two variables have the same value. In order to frame the graph edit distance within this framework, Riesen and Bunke propose a cost matrix constructed in such a way that graphs of different sizes can be compared, and where the cost of mapping vertices of one graph (the ``variables'') to vertices of the other graph (the ``values'') takes into account the labels of the vertices in question, and also the edges coming in and out of those vertices. In other words, this approximation considers only the {\em local} structure around each vertex in order to find the best mapping from $g_1$ to $g_2$, rather than the {\em global} structure. Experimental results show significant reduction of computation time with only a small performance penalty. Other approximation methods exist, as surveyed by \cite{gao2010survey}.

If the data of interest can be represented as trees, more efficient algorithms for tree data exist to calculate the {\bf tree edit distance}. When trees are ordered, the problem becomes tractable (polynomial complexity)~\citep{tai1979tree}, but it remains NP-complete for unordered trees~\citep{zhang1989editing}. Many polynomial algorithms exist for the case of ordered trees, such as that of~\cite{klein1998computing}. The reader is referred to the comprehensive overview by~\cite{bille2005survey}, for a complete list of approaches.

Additionally, the idea of graph edit distances has been employed to define similarity between other graph-related structures such as {\em processes}. For example, the work of \cite{montani2015knowledge} combines domain knowledge (to define the edit costs between different types of vertices) with graph edit distances to define a similarity function between processes (represented as graphs by having the different steps in a process represented as vertices, and the dependencies between these steps as edges, with some control structures, such as loops, also often represented as vertices).

Graph edit distances require setting, in advance, the edit operation costs. While this can be done manually, recent work from the field of {\em metric learning}~\citep{yang2006distance}. Metric learning focuses on the problem of learning a distance or similarity function given a training set. In the most common setting, a labeled training set of feature-vector instances is provided, and the problem is to learn a metric (typically a Mahalanobis distance) that is minimized for pairs instances with the same label, and maximized for pairs of instances with different labels. While most metric learning work has focused on feature-vector representations, some work exists on structured representations. Many of these approaches (e.g. the work of \citep{neuhaus2007automatic}) are based on the {\em expectation-maximization} (EM) algorithm~\citep{dempster1977maximum}, and, although they can be used for trees, become intractable for general graphs~\citep{bellet2013survey}. However, some relatively recent work has started to produce practical approaches to learn metrics for graph data. For example, {\bf Good Edit Similarity Learning} (GESL)~\citep{bellet2012good} learns edit costs in the following way. Given a training set consisting of graphs with different labels, it first precomputes the number of the different types of edit operations (insertion, deletions, substitutions) required to match each pair of graphs in the training set. Then, an optimization process optimizes a cost matrix based on these numbers to maximally separate graphs with different labels, and keep graphs with the same labels close together. In this way, although the learned cost matrix might not be the optimal, there is no need to recalculate edit distances during the optimization process, as previous approaches required.

\subsection{Refinement Graph-based Functions}\label{subsec:graph-refinement}

Most structured distance and similarity functions described in this paper are specific to a given representation formalism (i.e., distances for Horn clauses cannot be used for labeled graphs or viceversa). {\em Refinement operators}, however, have been proposed as a way to define distance functions that apply to a large set of structured representations.

The key idea is to abstract away from the underlying representation, and assume just the existence of a few constructs: 
\begin{itemize}
\item {\em Subsumption relation}\footnote{Notice that in the description logics notation, subsumption is written in the reverse order since it is seen as ``set inclusion'' of their interpretations. Here, $x_1 \sqsubseteq x_2$ means that $x_1$ is more general than $x_2$, while in description logics it has the opposite meaning.}: given two structured instances $x_1$ and $x_2$, we say that $x_1$ {\em subsumes} $x_2$ (written $x_1 \sqsubseteq x_2$) if $x_1$ is {\em more general} than $x_2$. For example, in the case of graphs, subsumption could be defined as checking if $x_1$ is a subgraph of $x_2$.
\item {\em Refinement operator}: a downward refinement operator is a function $\rho$ that, given an instance $x$, generates other instances (refinements) that are more specific than $x$, i.e. instances that are subsumed by $x$~\citep{laag1998completeness}. A refinement operator is {\em locally finite} when it generates a finite amount of refinements; it is {\em complete} if all the instances that are more specific than $x$ can be generated by iterated refinement of $x$, and {\em proper} if $x \not\in \rho(x)$.
\end{itemize}

In our previous work~\citep{ontanon2016refinement}, we defined a collection of subsumption relations for labeled graphs with different semantics, and their corresponding refinement operators. The base subsumption relation was defined as ``$g_1$ subsumes $g_2$ if a subgraph of $g_2$ is isomorphic to $g_1$''. The refinement operator basically takes in a graph and generates all the possible graphs that can be formed by adding one more vertex or edge and assigning them a new label. In case the labels are organized in a hierarchy, refinement operators that can specialize the labels in the graph were also defined. These two constructs define what is known as the {\em refinement graph}, a directed graph where each vertex is a graph, and where edges represent refinement. The refinement graph is a semi-lattice, with a special element $g_\bot$ which is the graph with no edges and no vertices, and all other graphs can be generated by iterative refinement starting from $g_\bot$. Therefore, we can now see the problem of assessing distance or similarity between graphs as that of assessing similarity between elements in a hierarchy, and use all the measures described in Section \ref{subsec:hierarchies}, among others. For example:

\begin{itemize}
\item {\bf Antiunification-based similarity} ($S_\lambda$): given two graphs $g_1$ and $g_2$, it calculates their most specific ancestor in the refinement graph (their {\em anti-unifier}, $g_1 \sqcap g_2$), which is equivalent to the MCS if subsumption is defined as graph-isomorphism, and assesses similarity as:
\[S_\lambda(g_1, g_2) = \frac{|g_\top \xrightarrow{\rho} (g_1 \sqcap g_2)|}
					         {|g_\top \xrightarrow{\rho} (g_1 \sqcap g_2)| + 
					          |(g_1 \sqcap g_2) \xrightarrow{\rho} g_1| + 
					          |(g_1 \sqcap g_2) \xrightarrow{\rho} g_2|}\]
\noindent where $|g_1 \xrightarrow{\rho} g_2|$ represents the length of a refinement path that starts in $g_1$ and goes to $g_2$ by repeated application of the refinement operator $\rho$. Notice that the number of refinement steps necessary from $g_\bot$ to a graph $g$ can be seen as a measure of {\em size}, and thus, this measure is equivalent to the one presented by \cite{wallis2001graph} (described above) if subsumption is defined as graph-isomorphism, since the denominator is basically the size of the union graph.

\item {\bf Property-based similarity} ($S_\pi$): A major issue with $S_\lambda$ is that it is computationally impractical, except for very small graphs (as expected, since the MCS calculation can be seen as a special case of it). The key idea of the {\em property-based similarity} measure is to decompose each graph into a collection of smaller graphs (called {\em properties}), and then count how many of these properties are shared between two given graphs. The key advantages of this similarity function are that: (1) the re-representation of graphs into sets of properties (which is the expensive operation) only needs to be done once, and after that, assessing similarity has a lower computational cost, and (2) each of these properties can be seen as a {\em feature}, and thus, feature weighting methods can be applied in order to improve accuracy in the context of machine learning methods. Decomposing a graph into a collection of {\em properties} is done via an operation called {\em disintegration}~\citep{ontanon2012similarity}, which, depending on the structure of the refinement graph, ensures that we can reconstruct the original graph by {\em integrating} all the properties again into a single graph using the {\em unification} operation. Once the graph $g_1$ and $g_2$ have been disintegrated, into a set of properties $D(g_1)$ and $D(g_2)$ respectively, similarity is defined as: 
\[S_\pi(g_1, g_2) = \frac{|\{\pi \in P \,\,\, | \,\,\, \pi \sqsubseteq g_1 \,\,\, \wedge \,\,\, \pi \sqsubseteq g_2\}|}
						 {|P|}\]
\noindent where $P = D(g_1) \cup D(g_2)$. Moreover, a weighted version of this similarity function ($S_{w\pi}$) can be defined if a weight is defined for each property, and instead of counting the number of shared properties, we add their weights.
\end{itemize}

Refinement operator-based distance functions are related to hierarchy-based distance functions, as well as to MCS-based functions as described above. However, they are also very related to edit distances. A refinement operator can be seen as a function that generates new graphs by performing {\em edits} on it. A {\em downward} refinement operator only generates graphs that are more specific. The complementary concept of an {\em upwards} refinement operator generates graphs that are more general. Thus, by combining upward with downward refinement operators, we can generate the complete set of edit operations required for defining an edit distance. Since upwards refinement operators are basically the inversion of downward refinement operators, we could define the edit distance between two graphs as $|(g_1 \sqcap g_2) \xrightarrow{\rho} g_1| + |(g_1 \sqcap g_2) \xrightarrow{\rho} g_2|$, with obvious connections to $S_\lambda$. Moreover, as described in our previous work~\citep{ontanon2012similarity}, $S_\pi$ and $S_\lambda$ are equivalent if the refinement graph satisfies certain properties. 

In summary, refinement operator-based distance functions can be seen as a way to use ideas from distance functions for hierarchies to define those for graphs by means of the intermediate concept of the {\em refinement graph}.

\subsection{Graph Kernels}\label{subsec:graph-kernels}

Kernel methods, and {\em support vector machines}~\citep{hearst1998support} in particular, rose a few decades ago as a powerful family of machine learning methods that could be applied to a larte type of representation formalisms, given an appropriate kernel exists. The key idea behind these methods is that the core optimization processes required for performing classification, regression or even clustering can be formulated in terms of {\em inner products} (e.g., the usual dot product, when we are talking about Euclidean spaces). Given data in some representation formalism, e.g. graphs, we could define machine learning algorithms by first transforming this data into some feature-vector representation with some mapping function $\phi$ and then operating using inner products over this feature vector representation. A {\em kernel function} $k$ is a function that given two data points $x_1$ and $x_2$ in some representation formalism, calculates the result of mapping these data points to a implicit feature-vector space and then calculating their inner product: $k(x_1,x_2) = \langle \phi(x_1), \phi(x_2)\rangle$ (where $\langle \cdot , \cdot \rangle$ represents the inner product). In this way, given a proper kernel function, the same learning algorithm can be applied to graph data, feature vector data, tree data, etc.

Kernel functions can be seen as similarity functions (since, the more similar two data points are, the higher their inner product). However, kernels must satisfy the property of being {\em positive definite}, which intuitively means that, for a given kernel, a finite or infinite feature-vector space must exist such that the kernel is equivalent to transforming the data to this space and then calculating an inner product in this space (the reader is referred to existing overviews of kernel methods for a formal definition of kernel functions~\citep{gartner2003survey,ralaivola2005graph} for a formal definition of kernels). Therefore, while all kernel functions can be seen as similarity functions, not all similarity functions are kernel functions. Thus, a significant amount of work exists in defining kernel functions that encapsulate or approximate edit distances, and other of the distance and similarity functions described above.

Graph kernels can be classified along many different axis. \cite{gartner2003survey} differentiate {\em model-driven} from {\em syntax-driven} kernels, and \cite{ralaivola2005graph} distinguish between adjacency matrix-kernels, marginalized graph kernels, and others. For the purposes of this paper, we will classify them by whether they apply to general graphs or to trees, and introduce the key ideas behind most kernels in the literature:

\begin{itemize}
\item {\em Tree kernels}: the two most common ideas for defining tree kernels are. 
	\begin{itemize}
	\item {\bf Tree traversal kernels}~\citep{smola2003fast}: the key idea is to transform a tree into a string by using a depth-first traversal of the tree. If the tree is unordered, we can assume a lexicographical order on the labels of the tree vertices and use it to define the tree traversal. After that, string kernels can be used to compare trees. Assuming the trees are not too unbalanced, tree traversal kernels are $O(n)$ (where $n$ is the number of vertices of the trees).
	\item {\bf Subtree occurrence kernels}: these are a particular type of {\em convolution kernels}~\citep{haussler1999convolution} (where object are divided into parts, and kernels are defied over these parts) applied to trees. For example \cite{collins2002convolution} propose a kernel based on counting how many subtrees two given trees share, and propose an efficient way to calculate this, as follows. Given a set of possible subtrees $T$, the kernel function for two trees $t_1 = \langle V_1, E_1, l_1 \rangle$, and $t_2 = \langle V_2, E_2, l_2 \rangle$ is defined as:
	\[k_{cd}(t_1, t_2) = \sum_{v_1 \in V_1} \sum_{v_2 \in V_2} C(v_1, v_2)\]
	\noindent where $C(v_1,v_2) = \sum_{t \in T} I_t(v_1)I_t(v_2)$, and $I_t(v_1) = 1$ if the subtree rooted at $v_1$ is identical to $t$, and $0$, otherwise. So, basically, $C(v_1,v_2)$ is the number of common subtrees of $t$ that can be found rooted both at $v_1$ and $v_2$.	This kernel, however, has the limitation that it can only be applied to trees where children of a vertex are distinguishable. Extensions of this kernel to lift this limitation were introduced by~\cite{kashima2002kernels}.
	\end{itemize}

\item {\em Graph kernels}: many different types of graph kernels have been proposed in the literature. However, most of them follow one of the following ideas.
	\begin{itemize}
	\item {\bf Subgraph occurrence kernels}: like subtree occurrence kernels, the key idea is to define a latent feature space consisting of all possible graphs, and then define the kernel function between two graphs $g_1$ and $g_2$, based on how many of those graphs do they share. \cite{gartner2003graph} showed that this computation is NP-hard for the general case of labeled graphs. Since subgraph occurrence kernels are computationally unfeasible, rather than considering all possible subgraphs, approaches that consider only certain types of structures, such as trees or cycles have been proposed~\citep{horvath2004cyclic}. A particularly common type of such types of approaches are {\em marginalized kernels}, based on ``random walks'', described below.
		
	\item {\bf Marginalized Kernels}: marginalized kernels are also a particular type of {\em convolution kernels} that derive from marginalized sequence kernels \citep{tsuda2002marginalized}. The key idea is to count the number of labeled walks two graphs share. Thus, the underlying infinite feature space is the set of all possible label sequences of length between 1 and $\infty$. Given two graphs $g_1 = \langle V_1, E_1, l_1 \rangle$, and $g_2 = \langle V_2, E_2, l_2 \rangle$, the basic formulation~\citep{ralaivola2005graph} of this kernel is as follows:
	
	\[k_m(g_1, g_2) = \sum_{i = 1}^{\infty} \sum_{\mathbf{s}^1 \in S_1^i, 
									     \mathbf{s}^2 \in S_2^i} k_{label}(l_1(\mathbf{s}^1), l_2(\mathbf{s}^2)) p(\mathbf{s}^1|g_1) p(\mathbf{s}^2|g_2) \]
	\noindent where $S_j^i$ is the set of all possible vertex sequences of length $i$ in graph $g_j$, $l_j(\mathbf{s})$ is the sequence of labels of a given vertex sequence $\mathbf{s}$ in graph $g_j$, $p(\mathbf{s}^j|g_j)$ is the probability of such vertex sequence given a user-defined transition probability function, and $k_{label}$ is a kernel between label sequences. \cite{Kashima2003kernel} proposed an efficient way to calculate this kernel via solving a set of simultaneous linear equations. Several enhancements to the basic kernel by Kashima et al. have been proposed in the literature such as enhancements for graphs (such as those appearing in chemistry) with lots of repeated labels, or removing the possibility of paths that ``go back'' on themselves~\citep{mahe2005graph}. Many other graph kernels based on random walks exist, such as the recent work by \cite{zhang2018retgk} based on the idea of the {\em return probability} of a random walk. Finally, as pointed out by several authors~\citep{tsuda2002marginalized,ralaivola2005graph} some other common types of kernels, such as Fisher kernels~\citep{jaakkola1999exploiting}, are particular cases of marginalized kernels. 
		
	\item {\bf Fingerprint kernels}: two types of kernels are referred to as {\em fingerprint kernels} in the literature. {\bf Traditional fingerprints} are commonly used in chemoinformatics and consist of bit vectors, where each bit corresponds to a chemical substructure (the list of chemical substructures to consider is usually set by hand using scientific literature on chemistry). The fingerprint of a molecule is calculated by setting to 1 all the bits corresponding to the substructures that the given molecule contains. Kernels are then just the inner standard product in the fingerprint vector space (notice that this is basically the same ideas as a subgraph occurrence kernel, but considering only a curated predefined set of subgraphs). On the other hand, {\bf hashed fingerprints} are a rather different type of kernel, where there is no predefined set of chemical structures. Instead, given a graph $g$ (usually representing a molecule), all possible paths starting from each vertex are computed, and for each path the corresponding label sequence (with the labels of all the vertices and maybe also edges traversed by the path) is determined. Then, each sequence of labels is used to calculate a has value $v$, used to generate a fixed sequence of bits. The final fingerprint of the graph is calculated as the bit-wise {\em OR} operation between al the bit sequences for each path. \cite{ralaivola2005graph} present an efficient way to calculate these fingerprints when considering all the paths from length 1 to infinity, and use it to define three three kernels based on these bit vectors (the {\bf Tanimoto kernel}, the {\bf MinMax kernel} and a {\bf Hybrid kernel} that is just a linear combination of the previous two). For example, the Tanimoto kernel is defined as:
	\[k_{\mathit{tm}}(g_1, g_2) = \frac{k(fp(g_1), fp(g_2))}{k(fp(g_1), fp(g_1)) + k(fp(g_2), fp(g_2)) - k(fp(g_1), fp(g_2))}\]
	\noindent where $fp$ is the bit vector with the fingerprint of a given graph, and $k$ is a regular kernel between vectors.
	
	\item {\bf Edit-distance kernels}: although the edit distance between two graphs does not satisfy the necessary conditions to define a kernel, several kernels have been proposed inspired by edit distances. For example, \cite{neuhaus2006edit} present a ``pseudo-kernel'' (since the resulting function is not guaranteed to be a kernel), based on selecting a reference graph $g_0$ and calculating the kernel function for two graphs as a function of their edit distances with respect to $g_0$. Although interesting, the main issue of this function is that it cannot be guaranteed to be positive definite. Another approach by the same authors is the {\bf convolution edit kernel}~\citep{neuhaus2006convolution} which is guaranteed to be positive definite, and is defined as follows. Let us assume that given a graph, we impose some arbitrary order over its vertices. Now, each graph is represented as a sequence of vertices. Given two graphs and their sequence of vertices, if we consider two subsequences (one from each graph) of the same length, they can be seen as defining a {\em mapping} between vertices of the two graphs (where the first vertex of the first subsequence is mapped to the first vertex of the other graph, and so on). We can not define the kernel as:
	\[k_{cek}(g_1, g_2) = \sum_{x_1 \in R(g_1), x_2 \in R(g_2)} k_{val}(x_1, x_2) \prod_{i = 1, ..., \mathit{length(x_1)}}k_{\mathit{subst}}(x_1[i], x_2[i])\]
	\noindent where $R$ is the set of all possible subsequences of vertices of a graph, $k_{val}$ is 1 is the subsequences $x_1$ and $x_2$ are the same length and 0 otherwise, and $k_{\mathit{subst}}$ is the substitution cost of substituting a vertex in one graph by a vertex in the other. While this is not equivalent to a full edit distance, it is a reasonable approximation in order to satisfy positive definiteness, and which has been shown experimentally to perform better (when used in a support vector machine) than a traditional edit distance in a $k$-nearest neighbor framework~\citep{neuhaus2006convolution} for some image and character recognition tasks.
	\end{itemize}
\end{itemize}

The list above captures some of the historically most common ideas used in graph kernels. However, Many other graph kernels have been proposed in the literature, such as those based on adjacency matrices~\citep{gartner2003graph}, among others. For example, recent work has proposed embedding each graph vertex using the adjacency matrix, and then using a distance metric, such as the earth-moved distance described above between the resulting embeddings~\citep{luss2008support}. This does not result in a positive definite kernel, but can be combined with indefinite kernel SVM methods~\citep{luss2008support} to achieve state of the art performance. 

Another recent idea is that of using the idea is that of using the $k$-core decomposition of a graph~\citep{nikolentzos2018degeneracy}, which decomposes a graph $g$ into a series of nested graphs: $g \supseteq c_0 \supseteq c_1, ...$, where $c_k$ is the $k$-core of $g$ (a largest subgraph of $g$ where all vertices have at least $k$ edges). This $k$-core decomposition captures the structure of a graph at different levels of granularity. Thus, the idea is to assess similarity between graphs at different granularities, since graphs might exhibit different structures at different levels of granularities.


\subsection{Graph Neural Networks}\label{subsec:graphnn}

A recent approach to assess similarity between graphs focuses on using {\bf graph neural networks} (GNNs). A GNN is a particular type of neural network capable of learning representations of graphs or vertices from graphs and that can be used for many supervised learning problems with graph data~\citep{battaglia2018relational}. Specifically, in order to use them for similarity function learning, GNNs have been used to embed graphs into vector space. This embedding is learned end-to-end in a supervised learning fashion, given a training set of graphs with annotations of which should be considered similar and which should be considered dissimilar. The resulting neural networks are called {\bf graph matching networks}~\citep{li2019graph}.

Two advantages of this approach are: (1) the embedding is learned directly from data, and thus the resulting similarity function is fitted to the task at hand similar to metric learning methods (see Section \ref{subsec:metriclearning}); and (2) once the graph embedding has been learned, similarity is computed only in the vector space, thus allowing for efficient retrieval techniques.

\subsection{Graph Vertices}

Finally, a very different family of distance or similarity functions concern comparing vertices within a graph. The problem of comparing vertices in a graph arises naturally when we think of graphs representing web pages (with edges representing links), or academic publications (with edges representing citations). These functions are very different from all the functions presented above, since the data being compared is itself not a graph, but rather {\em lies within} a graph. 
The underlying assumption of this line of work is that we do not have access to a set of features describing the vertices to be compared, and we need to compare them based on the graph structure. We will discuss some of the most common functions here for completeness, and refer the reader to Section 3 of the overview by \cite{lu2011link} for a more comprehensive list.

Many similarity functions have been proposed between graph vertices, which can be roughly classified into {\em local} vs {\em global} functions depending on whether they utilize only information concerning the immediate neighborhood of a vertex, or if they utilize the whole graph structure in order to calculate similarity. 

Given two vertices $v_1$ and $v_2$,  {\bf local similarity functions} between graph vertices are usually defined by assessing similarity between the neighborhood sets $\Gamma(v_1)$ and $\Gamma(v_2)$, containing all the vertices that are connected via a direct edge to $v_1$ and $v_2$ respectively. Given these two sets, vertex similarity is then usually assessed via the use of set similarity functions (like the Jaccard index, or the S{\o}rensen's Index described above). Early work in this direction can be traced back to the early work of \cite{small1973co}, who proposed the idea of {\bf co-citation} as a means to measure the relationship between two scientific documents. The {\em co-citation} index between two documents $v_1$ and $v_2$ is the number of documents that contain cites to both $v_1$ and $v_2$. Assuming both $v_1$ and $v_2$ are vertices on a graph $g_1 = \langle V, E, l \rangle$:
\[s_{\mathit{co-citation}}(v_1, v_2) = |\{v \in V | (v,v_1) \in E \,\, \wedge (v,v_2) \in E\}|\]
Notice that co-citation is basically measuring the size of the intersection of the directed neighborhoods of two vertices.

In contrast, {\bf global similarity functions} between graph vertices are defined using global properties of a graph. For example, most of them use the concept of {\em paths} in the graph. An early example of these functions is the {\bf Katz} index~\cite{katz1953new}, which counts the number of paths of different lengths that connect two given vertices, using a decay function on the length of these paths:
\[s_{\mathit{Katz}}(v_1, v_2) = \sum_{l=1}^{\infty}\beta^l |paths_{v_1 \to v_2}^l|\]
\noindent where $0 < \beta < 1$ is a decay constant, and $paths_{v_1 \to v_2}^l$ is the set of all possible paths from $v_1$ to $v_2$ of length $l$.

More recent work includes the {\bf SimRank} algorithm~\citep{jeh2002simrank} (called SimRank for its underlying similarity with PageRank~\citep{page1999pagerank}). SimRank assesses similarity between vertices based on the idea that vertices with similar connections (edges) are similar. The basic recursive formulation of SimRank is as follows:
	\[sim(v_1,v_2) = \frac{C}{|I(v_1)||I(v_2)|}\sum_{i=1}^{|I(v_1)|}\sum_{j=1}^{|I(v_2)|}sim(I_i(v_1), I_j(v_2))\]
\noindent where: $I(v)$ is the set of {\em in-neighbors} (vertices with an edge pointing to $v$), $C$ is a constant between 0 and 1, and $sim(v_1,v_2) = 1$ when $v=w$, and $sim(v_1,v_2) = 0$ if $|I(v_1)||I(v_2)| = 0$. 

SimRank can be interpreted as the probability that two random walkers starting at the two nodes in question would meet if walking the graph backwards~\citep{jeh2002simrank}. This idea of random walks, has been explored in several other similarity function. For examples \cite{pons2005computing} proposed the following distance function between two vertices $v_1$ and $v_2$ in a graph:

\[d_{PL}(v_1, v_2, t) = \sqrt{\sum_{k=1}^{n} \frac{(P_{1\to k}^t - P_{2\to k}^t)^2}{|I(v_k)|}}\]

\noindent where $|I(v_k)|$ is the number of incoming edges in $v_k$, $t$ is a parameter of the distance determining the length of the random walks, and $P_{i \to j}^t$ is the probability that a random walk of length $t$ starting in $v_i$ ends in $v_j$. The idea is that if two vertices belong to the same neighborhood in a graph (and should thus be considered similar), the probabilities of reaching all the other vertices in the graph should be similar. Pons and Latapy then proposed efficient ways to approximate such distance and used then to define an algorithm called {\em Walktrap} to identify the different ``communities'' (or clusters) of vertices in a graph in a computationally efficient way.


\section{Distance Functions for Logic-based Representations}\label{sec:logic-distance}

Research on distance functions for logic representations has occurred fairly independently in different communities, each focusing on a different logical formalism, with little interaction. Specifically, the three representation formalisms that have received more attention are Horn clauses, description logics and feature terms. Moreover, even if work has been carried out independently, many of the key underlying ideas are shared across these different pieces of work. 

Logical representations distinguish between {\em syntax} and {\em semantics} (given a target {\em domain}, the syntax defines the rules that determine which logical expressions can be written in a given logical formalism, and semantics determines the sets of individuals in the target domain that are covered by the different logical expressions). Thus, work exists on distance measures between logical expressions (clauses) and also between individuals. However, work on similarity between clauses is the most common (and most work on similarity between individuals actually first calculates what is known as the {\em most specific concept}, the clause the most closely represents an individual, and then uses distance between clauses). 

Most distance functions between logical clauses can be classified in two broad categories: {\em syntactic} (or {\em intensional}), and {\em semantic} (or {\em extensional}). The former are based on comparing the syntactic descriptions of logical clauses, and the latter are based on comparing the sets of individuals covered by the logical descriptions. Additionally, some distance functions combine ideas of both. Finally, there has also been work on trying to capture some of these ideas of distance and similarity as kernels, which we will cover at the end of this section.

The key difference between logic-based representations and graph-based representations is that logic-based representations afford inference processes to be performed over instances. For example, given an instance described as a logical clause, if background knowledge is available, additional facts about the instance can be potentially inferred. Thus, even if it's always possible to take a logical clause and represent it as a graph (having constants and functors be the vertices, and using edges to represent which functors and constants are the parameters of which other functors), this transformation loses the ability to perform inference. Thus, additional desirable properties have been proposed in the literature for similarity functions for logical representations. Below, we provide formal definitions of the three properties informally proposed by~\cite{dAmatoSF08}. Let $I$ be the {\em interpretation} function that defines the semantics of a given logic formalism (that maps logical clauses to the sets of individuals covered by them), let $x_1$, $x_2$ and $x_3$ be three clauses, $d$ be a distance function, and $s$ a similarity function.

\begin{enumerate}
\item {\em Soundness}: if $(I(x_1) \cap I(x_3)) \subseteq (I(x_2) \cap I(x_3))$ then $d(x_1,x_3) \geq d(x_2, x_3)$.

Intuitively, this means that if all the individuals covered by $x_1$ and $x_3$ are also covered by $x_2$, but that $x_2$ covers some additional individuals also covered by $x_3$, then $x_2$ is semantically closer to $x_3$, and thus the distance between $x_2$ and $x_3$ should be lower than that between $x_1$ and $x_3$. Analogously, $s(x_1,x_3) \leq s(x_2, x_3)$

\item {\em Equivalence Soundness}:  if $I(x_1) = I(x_2)$ then $d(x_1,x_3) = d(x_2, x_3)$. And, of course $s(x_1,x_3) = s(x_2, x_3)$.

Intuitively, if $x_1$ and $x_2$ are semantically equivalent given the logic at hand (i.e. their set of interpretations is the same), then the similarity between $x_1$ and $x_2$ to any other instance should be equal.

\item {\em Disjointness incompatibility}: imagine that $I(x_1) \cap I(x_3) = \emptyset$ and that $I(x_2) \cap I(x_3) = \emptyset$, all distance functions based on semantics will assess the distance between $x_1$ and $x_3$ and between $x_2$ and $x_3$ to be maximal, since their interpretations are disjoint, i.e., there is no individual that is covered at the same time by $x_1$ and by $x_3$. However, consider the following example: $x_1$ represents flights coming out of Berlin going to Frankfurt, $x_2$ flights coming out of Barcelona going to Philadelphia, and $x_3$ flights coming out of London going to Philadelphia. Clearly, their interpretations are disjoint, but $x_2$ and $x_3$ share the fact that flights go to the same destination. Distance functions that are able to capture this similarity even when the interpretations of the two clauses are disjoint are said to be able to handle {\em disjointness incompatibility}.
\end{enumerate}

Let us now summarize the existing work on distance and similarity functions for logical representation formalisms in view of these new properties, and also compared to the work presented before for graph-based representations.

\subsection{Syntactic Distance Functions}\label{sec:logic-syntactic}

Syntactic distance functions compare instances by directly comparing the logical expression used to represent them. Let us classify the work based on the logical representation formalism used.

\subsubsection{Horn Clauses}
An early representative method of this idea is that of \cite{hutchinson1997metrics}, who studied metrics between logical {\em terms} and logical {\em clauses}. 

Given an alphabet of variables $X$, a an alphabet of function symbols $F$, a term is either a variable in $X$, or an expression of the form $f(t_1, ..., t_n)$, where $f \in F$ and $t_1, ..., t_n$ are terms (a constant is just a term or zero arity). {\bf Hutchinson} proposed to measure the distance between two terms by using the ideas of {\em variable substitutions} and {\em least general generalizations} ({\em lgg}). Given two terms, $t_1$ and $t_2$, and their {\em lgg}, $t^*$, let $\theta_1$ and $\theta_2$ be the variable substitutions that turn $t^*$ into $t_1$ and $t^*$ into $t_2$ respectively. The distance between two terms is then defined as:

\[d_{H}(t_1, t_2) = |\theta_1| + |\theta_2|\]

\noindent where $| \cdot |$ is some size function on variable substitutions (e.g., the number of variables being substituted). This idea can be extended to literals (a literal is a term that can be negated) by considering the {\em negation} symbol to be just a regular function symbol. And then, to clauses by considering that clauses are just sets of literals and then using the {\em Hausdorff distance}. Thus, given two clauses $C_1$ and $C_2$, their distance can be assessed as:

\[d_{H}(C_1, V_2) = max\left(\max_{t_1 \in C_1} \min_{t_2 \in C_2} d_H(t_1,t_2), \max_{t_2 \in C_2} \min_{t_1 \in C_1} d_H(t_1,t_2)\right) \]

Terms and clauses often refer to individuals, e.g. the term {\em mother(alice, bob)}, intuitively states that the individual named {\em alice} is the mother of the individual named {\em bob}. So, it is often useful to assess the distance between individuals referred to by terms, rather than  the distance between terms themselves. Early work in this direction is the work of \cite{bisson1990kbg}. Consider a knowledge base consisting of a set of terms. Given an individual $x$, let $(f, n)$, where $f$ is a function symbol and $n$ is an integer, be an {\em occurrence} of $x$ if there is a term in the knowledge base with function symbol $f$ and where $x$ appears as the $n$-th argument. Let now $\mathit{occurrences}(x)$ be the set of all occurrences of an individual $x$ in the knowledge base. {\bf Bisson's similarity function} between individuals is defined as:

\[s_{B}(x_1, x_2) = \frac{|\mathit{occurrences}(x_1) \cap \mathit{occurrences}(x_2)|}{max(|\mathit{occurrences}(x_1)|, |\mathit{occurrences}(x_2)|)}\]

\noindent in other words, their similarity is defined as a pseudo-Jaccard index (replacing the size of the union in the denominator by the max size) of their sets of occurrences.

This work was later extended to account for similarity between the different occurrences~\citep{bisson1992similarity}. In this extension, the similarity between two entities (SIM) is calculated as the average of the similarity of the terms in their common occurrences (T-SIM). Incidentally, SIM depends on T-SIM, and T-SIM depends on SIM. So, this results on a system of equations that needs to be solved in order to assess the similarity between two entities. This system of equations is often non-linear, and thus Bisson proposed to use Jacobi's method~\citep{golub2012matrix} to solve it.

Probably one of the best known similarity functions for logic-based representations is in {\bf RIBL} (Relational Instance-Based Learning)~\citep{werner96relational}. RIBL's measure is a modification of Bison's similarity function~\citep{bisson1992similarity} so that rather than considering a network of predicates (thus requiring Jacobi's method to solve a system of equations), it builds a hierarchical representation in the form of a tree that is a string generalization of standard similarity functions for feature vectors. Specifically, this similarity function is defined for Horn-clause style representations (such as the one shown in Figure \ref{fig:representations}.e) and works as follows. Given two entities, each described by a logical term, where some of the attributes of the terms are primitive values (e.g., numbers), and some others are references to other objects, the similarity of the two entities is assessed as the similarity of their attribute's values. If some of these attributes are references to other objects, then their similarity is assessed recursively:

\[\text{sim-e}(f(t_{1,1}, ..., t_{1,m}), f(t_{2,1}, ..., t_{2,m})) = \frac{\sum^m_{i=1, i \in \text{Input-Args}(f)} \text{sim-a}^{type(f, i)}(t_{1,i}, t_{2,i}, 0)}{|\text{Input-Args}(f)| }\]

\noindent where $ \text{Input-Args}(f)$ is the subset of arguments of $f$ that are considered ``input arguments'' (RIBL distinguished between input and output arguments in predicates), $type(f,i)$ is the ``data type'' of the argument $i$ of functor $f$ (numeric, symbolic, reference to another object, etc.), and $\text{sim-a}^{type(f, i)}$ is a collection of functions (one per different data type of the arguments) that recursively assess the similarity of the arguments. Thus, notice that if all arguments are numeric or symbolic, this is basically a standard feature vector similarity function (the average similarity of all the attributes), but if any attribute is a reference, then $\text{sim-a}^{type(f, i)}$ will recursively call $\text{sim-e}$. The $0$ as the third parameter of $\text{sim-a}^{type(f, i)}$ refers to the depth at which we are doing recursive calls, since usually a maximum depth limit is set for RIBL, to prevent infinite recursion.

The basic idea of RIBL was extended in the work of \cite{Horvath2001RIBL2}, defining additional versions of $\text{sim-a}^{type(f, i)}$ that support arguments of type {\em list} or {\em term} using edit distances. Thus, notice that the key idea of RIBL is just to assess the similarity of predicates by the similarity of their attributes, which is then assessed recursively in case any attribute is in itself a reference to another entity, by assessing the similarity of the predicates describing those entities. This is a representative example the idea of {\bf hierarchical aggregation}, which many other distance functions we will describe below follow. Also, notice that RIBL requires specific similarity functions for every data type that is to be used in the definition of the logical predicates.

Other hierarchical aggregation measures include the work of \cite{nienhuys1997distance} where a distance function between ground atoms is presented, based on considering atoms to be trees, and using a hierarchical recursive definition. Then this distance is extended to clauses using the same idea of the Hausdorff distance used by Hutchinson as explained above. This work was extended by~\cite{ramon1998framework}, to allow for non-ground atoms.

\subsubsection{Feature Terms}\label{subsec:fterms-similarity}
Another framework for assessing similarity using the idea of hierarchical aggregation is the work of \cite{armengol2001similarity,armengol2002similarity} with their {\bf LAUD} and {\bf SHAUD} similarity functions. These functions focus on a logical formalism called {\em em feature terms}~\citep{Carpenter1992}. Specifically, SHAUD (which is an improvement over the previous LAUD similarity function), works as follows. Given two instances $c_1$ and $c_2$ represented as feature terms (see Figure \ref{fig:representations}.d for an example feature term), their similarity is defined as:

\[sim_{\mathit{SHAUD}}(c_1, c_2) = \frac{1}{r} \sum_{\langle s_i, w_i, r_i \rangle \in T(CS(c_1, c_2))} s_i * w_i\]

\noindent where $r$ is a normalization value to make the similarity take values between 0 and 1, $CS$ refers to the ``common structure'' between $c_1$ and $c_2$, i.e., the set of attributes that the roots of $c_1$ and $c_2$ share (for example, in the feature term in Figure \ref{fig:representations}.d, the common structure between $X_2$ and $X_3$ are the {\em length} and {\em shape} attributes). $T$ is a function that for each shared attribute $f$ computes a tuple $\langle s_i, w_i, r_i \rangle$, where $s_i$ is the similarity of  $c_1.f$ and $c_2.f$, and $w_i$ and $r_i$ are a measure of the ``size'' of $c_1.f$ and $c_2.f$: $w_i$ measures the number of variables in their shared structure (e.g., the size of their intersection) and $r_i$ measures the total size (i.e., the size of their union).

In order to calculate $s_i$, SHAUD, like RIBL, uses a hierarchical process, where if $c_1.f$ and $c_2.f$ are numerical or categorical values, special similarity functions are used, but if they are structured terms, the SHAUD similarity is called recursively.

As we noted in our previous work~\citep{ontanon2012similarity}, hierarchical aggregation methods like RIBL and SHAUD make two underlying assumptions: (1) that data is organized hierarchically in a tree form (for example, RIBL requires a maximum depth parameter to avoid infinite recursion in case data forms loops, and similarly SHAUD would get stuck in an infinite recursion with feature terms that contain cycles); (2) they implicitly assume that information that is ``deeper'' in the tree is less important than information that is found earlier in the tree, which is an arbitrary assumption in many real-world datasets.

\subsubsection{Description Logics}
A significant amount of work has been done on similarity functions for {\em Description Logics}~\citep{DL2003handbook}. Concerning syntactic functions, one of the earliest examples is the similarity function proposed by \cite{gonzalez1999applying}, where they proposed to assess the similarity between two individuals as the sum of the similarity between the most specific concepts of which those individuals are instances of, and the similarity of their {\em roles} (where ``role'' is the term used in Description Logics to refer to the concept of attributes or features of individuals). Specifically, the proposed similarity function between two individuals $x_1$ and $x_2$ is defined as:
 \[
 sim_{GC}(x_1,x_2) = \left\{
 \begin{array}{ll}
 sim(t(x_1), t(x_2)) 		& \text{if} \,\, \forall r \in R: \\
 					& x_1.r = x_2.r = \emptyset \\
 \frac{1}{2} \left(  sim(t(x_1), t(x_2)) + \frac{\sum_r \in R sim(x_1.r, x_2.r)}{|\{r \in R : x_1.r \neq \emptyset \wedge x_2.r \neq \emptyset \}|} \right) 	& \text{otherwise}
 \end{array}
 \right .
 \]
\noindent where $t(x)$ is the most specific concept of which an individual $x$ is an instance, $R$ is the set of all possible roles, and $x.r$ is the set of individuals connected to $x$ via role $r$. If $x.r$ is a set, and not just one individual, $sim(x_1.r, x_2.r)$ is defined by calculating the sum of the similarities between each individual of $x_1.r$ and the corresponding individual in $x_2.r$ with the maximum similarity. Notice that this definition might contain infinite loops. In order to prevent this, roles that cause circular cycles are not considered as part of the similarity calculations.

Another example is the work of \cite{janowicz2006sim}, who present a similarity framework called {\bf SIM-DL} for comparing $\mathcal{ALCNR}$ concept descriptions. Concept descriptions in $\mathcal{ALCNR}$ normal form are represented as disjunctions of other concepts. Given two concept definitions: $C = C_1 \sqcup ... \sqcup C_n$ and $D = D_1 \sqcup ... \sqcup D_m$, SIM-DL assesses their similarity as:
\[sim_u(C,D) = \sum_{(C_i, D_j) \in SI} w_{ij} \times sim_i(C_i, D_j)\]
\noindent where $sim_u$ stands for similarity between concepts described as the union of concepts, and $sim_i$ is a recursive call for concepts represented as the intersection of other concept definitions. Similarly, $sim_i$ recursively calls to $sim_p$ (between primitive concepts), and other functions for existential quantifier definitions, role definitions and value restrictions. Also, when comparing definitions between concepts in $sim_u$, SIM-DL first calculates the similarity between each pair in the Cartesian product of $C_1, ..., C_n$ and $D_1, ..., D_n$. Then, for each $C_i$, the corresponding $D_j$ with the highest similarity is selected. The {\em selected pairs} form the set $SI$. Finally, the weights $w_{ij}$ have to be set so that they add up to 1, but the authors leave the specific weighting function open, and just mention that they could be computed, for example out of the set cardinality of the individuals covered by each concept. Finally, given the non-symmetry of the step concerning the selection of pairs for $SI$, SIM-DL does not directly satisfy the {\em symmetry} property from Definition \ref{def:similarity}. However, notice that this is not a crucial property, since any non-symmetric similarity function can be turned into a symmetric one by calculating $(s(x,y) + s(y,x))/2$.

In summary, notice that SIM-DL is basically a recursive syntactic similarity function similar to the work of  \cite{gonzalez1999applying}, but working over concept definitions, rather than over individuals.

\subsection{Semantic Distance Functions}\label{sec:logic-semantic}

The key characteristic of semantic distance functions is that rather than using the syntactic representation of a concept to assess similarity, they assess similarity based on the set of individuals covered by concept definitions (i.e., their semantics). These measures are sometimes referred to as ``extensionality-based similarities''~\citep{dAmatoSF08}, as they are based on enumerating the set of individuals covered by a concept (their ``extension''). The basic idea behind these semantic or extensional measures is Resnik's idea of {\em information content} described in Section \ref{subsec:hierarchies}.

An early example of a semantic distance function can be found in the work of \cite{sebag1997distance}. Sebag proposed {\bf DISTILL}, one of the first distance functions that was not based on the syntax of the description of a given instance, but on inducing a collection of {\em discriminant hypotheses}. The idea is to pick random pairs of examples of different classes, and find hypotheses (concept descriptions) that separate them. After this, each instance is re-represented as a boolean vector (with one position per hypothesis, representing whether the instance satisfies the hypothesis or not). Distance between instances can then be defined as a Hamming distance between these vectors. 

Sebag's idea is related to what has later been referred to as {\em fingerprinting} similarity functions (see fingerprinting kernels described above in Section \ref{subsec:graph-kernels}), or as {\em binary hashing}~\citep{datar2004locality}, which are common in the literature of computational biology, and on information retrieval. Also, notice that this idea is also related to the idea of the {\em property-based similarity} described in Section \ref{subsec:graph-refinement}.

A significant amount of work on semantic distance functions has been carried out within the Description Logic community (see for example an early review by \cite{borgida2005towards}). An example of this line of work is the work of \cite{hu2006semantic}. They proposed the idea of {\bf unfolding concepts}, which means taking a concept definition and transforming it into a description that only contains ``primitive concepts'' from a Description Logic ontology via the application of a set of transformation rules (and forbidding circular concept definitions in the ontology to ensure termination). Once unfolded, concepts descriptions can be transformed into a {\em signature vector} with one position per primitive concept in the ontology, and where the value corresponds to the number of times that each concept appears in the unfolded concept definition. Computing distances between concepts is then reduced to computing the distance between these vectors. Specifically, they propose to calculate a weight for each vector position using {\em term frequency - inverse document frequency} (TF-IDF)~\citep{singhal2001modern}. One particularity of this distance function is that, in order to capture negation, they reverse the sign of the weights for concepts appearing with a negation in a concept definition, thus, using their proposed similarity function equation, some concepts might have negative similarity, which violates some of the basic properties of similarity and distance functions (see Definition~\ref{def:similarity}):

\[s_{Hu}(C_1, C_2) = \frac{\sum_{w_i \in C_1, w'_i \in C_2} w_i \times w'_i}{\sqrt{\sum_{w_i \in C_1} w_i^2}\sqrt{\sum_{w'_i \in C_2}{w'_i}^2}}\]

\noindent where $C_1$ and $C_2$ are two signature vectors representing two concepts, and $w_i$ represent the TF-IDF weights for each of the primitive concepts in the ontology for each of the two signature vectors (notice that these weights are negative if the primitive concept appears negated in the definition).

A hybrid measure that integrates syntactic and semantic information was proposed by \cite{d2006dissimilarity} to compare concepts in the $\mathcal{ALC}$ Description Logic. Specifically, they propose a distance function defined recursively (such as the syntactic measures described above), but that is employs a Resnik-style semantic measure to compare primitive concepts. For example, to compare two concept definitions $C = C_1 \sqcup ... \sqcup C_n$ and $D = D_1 \sqcup ... \sqcup D_m$ defined as the union of sets of more primitive concepts, the distance function is defined as follows (similar to the syntactic measures above):

\[
f_\sqcup(C, D) = 
\left\{
\begin{array}{ll}
0	& \text{if} \,\, C \equiv D	\\
\infty	& \text{if} \,\, C \sqcap D = \bot 	\\
max_{i \in [1,..,n], j \in [1, ..., m]} f_\sqcap(C_i, D_i)	& \text{otherwise} 	\\
\end{array}
\right.
\]

This definition then recursively calls $f_\sqcap$, etc. decomposing the distance function based on the different Description Logic constructs to define concepts. In the end, when comparing primitive concepts, the distance function is defined as:

\[
f_\mathit{primitive}(C, D) = 
\left\{
\begin{array}{ll}
\infty	& \text{if} \,\, C \sqcap D = \bot 	\\
\frac{IC(C \sqcap D) + 1}{IC(C \sqcup D) + 1}	& \text{otherwise} 	\\
\end{array}
\right.
\]

\noindent where $IC$ stands for {\em information content} and is assessed as $IC(C) = \log P(C)$), where $P(C)$ is the probability of encountering an instance of concept $C$, which is estimated using the individuals in the ABox. Thus, as can be seen, this function combines both syntactic and semantic elements to compare concept descriptions in Description Logic. The proposed approach can be extended to compare individuals by using the idea of the {\em MSC} (most specific concept), which is the most specific concept description that covers an individual. So, to compare two individuals, we compute their MSCs, and then assess the distance between them.

An interesting note is that semantic distance and similarity functions tend to violate the {\em disjointness incompatibility} property discussed above, whereas syntactic functions do not.

\subsection{Propositionalization}\label{sec:propositionalization}

Another traditional approach to apply machine learning methods in general in structured representations is that of {\em propositionalizarion}~\citep{kramer2001propositionalization, krogel2003comparative}. Propositionalization consists of translating a structured representation into a flat propositional (usually a Boolean fixed-size feature vector, but some approaches can create non-Boolean features), so that standard machine learning methods, or in our case distance functions for propositional data, can be applied. Propositionalization is related to the ideas of {\em predicate construction} or {\em predicate invention}~\citep{kok2007statistical}. 

A representative example of this approach is the SINUS system~\citep{krogel2003comparative}, which constructs features by systematically considering conjunctions of literals, and then evaluating them using a ``quality measure'' to filter out features that are not useful. 

Although propositionalization has not had widespread use for defining distance functions, it has been used implicitly for this purpose in the context of clustering. For example the COING system~\citep{bournaud2002propositionalization} clustered graph-based data by increasingly enlarging a propositional representation using propositionalization until a satisfactory clustering of the data has been reached.

\subsection{Refinement Graphs}

As mentioned above, distance functions defined over refinement graphs, are applicable to a large set of structured representations, given that appropriate refinement operators and subsumption relation are available. Both similarity functions described in Section \ref{subsec:graph-refinement} are applicable, and have been applied, to logic-based representations.

For example, \cite{ontanon2012similarity} defined refinement operators for feature terms and used them to define similarity functions, \cite{SanchezRuizGranadosOGP11} did the same for the $\mathcal{EL}$ description logic, and \cite{sanchez2016measuring} for description logic conjunctive queries. Moreover, refinement operators for other logical representations have been proposed in the literature, and can be used to define distance functions, for example: for $\mathcal{LC}$ description logic~\citep{LehmannH07}, $\mathcal{EL}$ description logic~\citep{LehmannH09}, $\mathcal{ALER}$ description logic~\citep{badea1999dlrefinement}, 

As with the distance functions based on refinement operators for graphs-based data, the main drawback of distance functions defined for logic-based representations is the computational complexity, as subsumption (required for distance calculations) tends to be an expensive operation. 

Distance functions defined based on refinement graphs could be considered as syntactic or as semantic depending on how the subsumption relation used is defined. If subsumption is defined over the syntax of the descriptions, then these are syntactic, and if it is defined over the interpretations of the descriptions, then these are semantic.

\subsection{Kernels for Logic-based Representations}\label{sec:logic-kernels}

Finally, there has also been a significant amount of work on defining kernels for logic-based representations, or encapsulating existing distance functions for logic-based representations into kernels. 

An early example of this line of research is the work of \cite{gartner2002kernels}, who defined a kernel for a typed higher-order logic based on an extension of Church's simple theory of types~\citep{church1940formulation} with type constructors, terms, and functions. The key idea is to assume the existence of a set of base kernels for the different data constructors of their logic representation. For example, for the data constructor $Nat$, representing the natural numbers, the {\em product kernel} ($k_{Nat}(m,n) = mn$) can be used. Then, given two terms $s = f_s(s_1, ..., s_n)$ and $t = f_t(t_1, ..., t_n)$, with functors of type $F$, the kernel is defined as follows:
\[k(t_1, t_2) = \left\{
\begin{array}{ll}
k_F(f_s, f_t)	&	\text{if} \,\, f_s \neq f_t \\
k_F(f_s, f_s) + \sum_{i = 1, ... n} k(s_i, t_i)		& \text{otherwise}
\end{array}
\right.\]
If $s$ and $t$ are two functions with type $S \to I$, then the kernel is defined as:
\[k(s,t) = \sum_{u \in supp(s), v \in supp(t)} k(s(u), t(v)) k(u,v)\]
\noindent where $supp(s)$ and $supp(t)$ represent the support of $s$ and $t$ respectively. Thus, notice that this kernel is basically another instance of the idea of hierarchical aggregation that was already present in distance functions such as RIBL or SHAUD, but in the form of a kernel.

{\bf SVILP} (Support Vector Inductive Logic Programming)~\citep{muggleton2005support} is a framework based on kernels for Horn clauses. The main difference with the kernel described in the previous paragraph is that the kernel in SVILP uses {\em logical background knowledge}. Thus, while \citeauthor{gartner2002kernels} kernel is syntactic and only considers the syntactic representation of terms, SVILP's kernel considers that there might be background knowledge $B$ in the form of logical rules, with which inferences can be drawn that affect the similarity calculations. Specifically, the kernel is defined as follows. Given a hypothesis space $\mathcal{H}$ (where every hypothesis is a logical clause), we say that a hypothesis $h \in H$ {\em covers} a specific instance $x$ if $B,h \vDash x$ (i.e., if the instance is entailed by the hypothesis and the background knowledge). Now, given a set of hypothesis $H \subseteq \mathcal{H}$, and a probability distribution over these hypotheses: $\pi : H \to [0,1]$ such that $\sum_{h \in H} \pi(h) = 1$, the kernel is defined as:
\[k(x_1, x_2) = f(\tau(x_1) \cap \tau(x_2))\]
\noindent where $\tau(x) = \{h \in H | B,h \vDash x\}$, and $f(H') = \sum_{h \in H'} \pi(h)$. Thus, the kernel is defined as the sum of the probabilities of the hypotheses that cover both instances (which can be shown to be a positive definite kernel).

Kernels have also been defined for Description Logics. For example, \cite{fanizzi2006declarative} defined a kernel for descriptions in the $\mathcal{ALC}$ Description Logic. The proposed kernel uses a very similar definition to the distance function by \cite{d2006dissimilarity}. In order to go beyond the kernel being a mere syntactic measure, they require concepts to be expressed in a {\em normal form}. Given this normal form, the kernel is then defined recursively depending on whether the top operator in the expressions is a disjunction, a conjunction or if we are down to the level of primitive concepts. Given two descriptions in normal form $D_1 = \sqcup_{i=1...n} C_i^1$ and $D_2 = \sqcup_{i=1...m} C_i^2$, the kernel is defined as:

\[k(D_1, D_2) = \lambda \sum_{i=1...n} \sum_{j=1...m} k(C^1_i, C^2_j)\]

\noindent where $\lambda \in (0,1]$ is used to lower the weight of comparisons done deep into the descriptions, and thus, it decreases with every recursive call of the kernel. At each recursive call, either the definition for disjunctions (shown above) or that for conjunctions is used, depending on the top operator of the descriptions, until reaching the level of primitive concepts, for which the set kernel defined by \cite{gartner2004kernels} is used to compare the interpretations of the primitive concepts (which are the sets of individuals from the ABox covered by the concepts). An extension of this kernel for the $\mathcal{ALCN}$ Description Logic was presented by \cite{fanizzi2008learning}.

Finally, another idea that has been used is to represent Description Logic expressions as graphs, and then use graph kernels. For example~\cite{de2015substructure} compared several of the kernels described in Section \ref{subsec:graph-kernels} such as subtree occurrence kernels, marginalized kernels, and compared them against simple bag of labels baseline kernels, showing that subtree occurrence kernels had the best performance.


\section{Distance Functions for Frame-based Representations}\label{sec:frame-distance}

Most work on frame-based or object-oriented representations has been inspired by the so called {\bf local-global principle}~\citep{wess1995fallbasiertes}, where similarity is assessed using two separate functions: a {\em local} similarity function is defined for individual properties or slots of the descriptions being compared, and a {\em global} similarity function is used to aggregate these local similarities. Notice that this idea is basically the same as the {\em hierarchical aggregation} idea described in Section \ref{sec:logic-syntactic}, and thus distance functions based on the local-global principle are based on the same ideas as most syntactic similarity functions between logical representations described above.

One of the best known local-global principle-based similarity function was presented by \cite{bergmann1998similarity}, dividing the similarity function calculation between two object-oriented representations in two steps: {\em intra-class similarity} and {\em inter-class similarity}). Intra-class similarity between two instances $x_1$ and $x_2$ is defined as:
\[s_\textit{intra}(x_1, x_2) = \Phi(s_\textit{local}(x_1.a_1, x_2, a_1), ..., s_\textit{local}(x_1.a_n, x_2, a_n))\]
\noindent where $\Phi$ is an aggregation function (e.g., the average, or the sum), $s_\mathit{local}$ is a similarity function between attribute values, and $a_1$, ..., $a_n$ are the shared attributes between the two instances. Inter-class similarity is assessed as:
\[s_\textit{inter}(x_1, x_2) = \left\{
\begin{array}{ll}
1 & \text{id} \,\, \mathit{class}(x_1) = \mathit{class}(x_2) \\
S_{\mathit{class}(x_1) \sqcap \mathit{class}(x_2)} & \text{otherwise}
\end{array}
\right.\]
\noindent where $\mathit{class}(x_1)$ represents the class of a given instance, and $\mathit{class}(x_1) \sqcap \mathit{class}(x_2)$ refers to the most specific common parent of the classes of both instances. In the work of Bergman and Stahl, they propose to annotate the class hierarchies with a similarity value $S_C$ for each class $C$. Similarity between two instances is then defined as $s(x_1, x_2) = s_\textit{intra}(x_1, x_2) * s_\textit{inter}(x_1, x_2)$.

Notice that, although not noted by the original authors, $s_\textit{inter}$ is basically a similarity function between elements in a hierarchy (Section \ref{subsec:hierarchies}), and thus, Rada's or Resnik's ideas can be used to define the $S_C$ values. Also notice that, as mentioned above, some logic-based similarity functions are very related to these ideas, and in particular, the LAUD similarity function mentioned in Section \ref{subsec:fterms-similarity} is a particular case of Bergman and Stahl's similarity function.

Several other similarity functions have been defined that follow the same idea. For example,  \cite{assali2009case} propose a similarity function that is a particular case Bergman and Stahl's, by defining $\Phi$ to be the average, and defining $S_C$ as:
\[s_C(x_1, x_2) = \frac{2 \times \mathit{depth}(\mathit{class}(x_1) \sqcap \mathit{class}(x_2))}{\mathit{depth}(\mathit{class}(x_1)) + \mathit{depth}(\mathit{class}(x_2))}\]
\noindent where $\mathit{depth}$ is a function that determines the depth of a given class in the class hierarchy (with the root node having depth 0).
Moreover, Assali et al. consider a framework where instances are represented as sets of descriptions (each of them an object-oriented description), and thus, to assess similarity they first need to find a mapping between descriptions of two instances, and then apply the equations above.

Additionally, even of similarity functions between workflows are better characterized as graph-based similarities, \cite{bergmann2014similarity} propose a measure that is a direct application of the local-global principle to comparing workflows. A workflow can be seen as a graph, where vertices represent processes, and edges represent control or data flow. In their framework they consider a graph-based representation of workflows where each edge and vertex is annotated with a Description Logic description. Thus, they use a local similarity function between edges and vertices, based on similarity functions for Description Logics, and a global similarity function based on finding a mapping between two workflows and then adding the similarity values of the pairs, normalized by the number of edges and vertices. As is well known from the graph matching literature (see above), finding this global mapping is intractable. The authors use an $A^*$ algorithm to calculate, but other modern graph matching algorithm could be used instead

Finally, the work on similarity functions for feature term representations above (Section \ref{subsec:fterms-similarity}) can be considered as distance functions for frame-based representations, since feature terms were conceived as a formalization of object oriented representations. Thus, measures such as LAUD, SHAUD or those based on refinement operators should also be considered to fit within this category.

\section{Discussion}\label{sec:discussion}

Sections \ref{sec:graph-distance}, \ref{sec:logic-distance}, and \ref{sec:frame-distance} have summarized existing work on distance and similarity functions for different structured representations. Although the literature on structured similarity assessment is vast, there are clear common themes that arise when looking at the body of work as a whole, which we will try to summarize in this section.


The first is that although the work has been classified along graph-based, logic-based and frame-based representations (with the purpose of providing structure to this paper), there is clear overlap between these areas. For example, frame-based representations are tightly coupled with logic-based ones. For example, the formalism called {\em feature terms} was precisely defined to provide a logical substrate to frame-based representations. As a matter of fact, frame-based similarity and distance functions are mostly based on ideas from syntactic similarity functions for logic-based frameworks such as ``hierarchical aggregation''.


A simple way to understand where does this overlap between the work on all three representations comes from is to analyze the basic underlying ideas that give rise to the different distance functions covered in this paper. Although
there is a very large number of distance functions proposed in the literature, they all stem from a small set of common ideas. Some of the most prevalent ones are:
\begin{itemize}
\item Quantify the amount of shared structure: ideas such as the Jaccard similarity, all the edge-counting functions, and those based on the calculation of the MCS or antiunification are instances of this idea. They are all based around determining the shared structure (MCS, antiunification, intersection, etc.), and then applying some metric to it to measure its size. Edit distances can also be seen as a variation of this idea (where the differences, rather than the similarities are counted), and as explained above, in some cases, it can be seen that edit distances and calculating shared structure (e.g., MCS) are equivalent.
\item Measure information content: this idea stems from the realization that not all shared structure is equally important. There might be shared features that are not relevant for the task at hand. Thus, information content-based measures use information theoretical measures to determine the amount of information that is shared between two structures. 
\item Fingerprinting: 
i.e., the idea of transforming a structured object into a (usually binary) vectors where each position corresponds to whether the object satisfies a certain test or not,
is another idea that appears in a significant amount of work, not just in kernel-based measures, but propositionalization techniques 
and some of the early semantic distance functions for logical structures can be seen as a particular instance of this.
\end{itemize}
Also, we should note that most of these ideas come from non-structured representations. For example, the idea of quantifying the  amount of shared structure can be seen as a generalization of the Jaccard-style similarity functions for sets to structured representations, 
and information content measures stem from distance functions between elements in a taxonomy.

Thus, many of the different functions covered in this paper can be seen as the different instantiations of these shared ideas, which take different forms in different representation formalisms. For example, while to measure similarity between sets using the Jaccard similarity, we need to calculate the ``intersection'' between sets, if we are doing so for graphs, we need to compute the MCS, and if we are doing so for logical expressions, then we need to calculate an antiunifier. 
Another example is the local-global distance functions for frame-based representations~\citep{wess1995fallbasiertes}, which are basically a direct generalization of Euclidean distances for vectors.

\begin{figure}[tb]
    \centering
    \includegraphics[width=\textwidth]{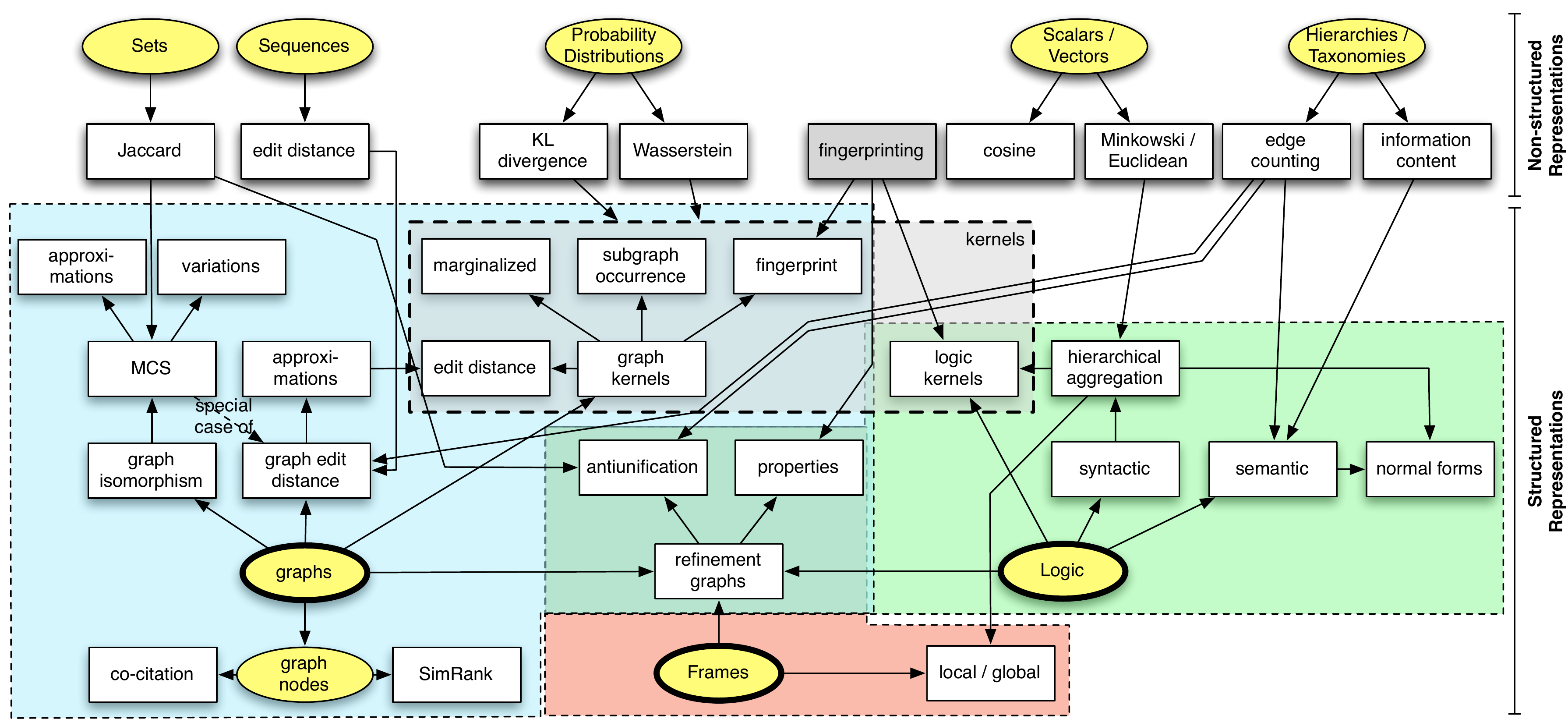}
    \caption{An overview of the main distance and similarity ideas reported in this paper, and how do they relate to each other.}
    \label{fig:overview}
\end{figure}

These ideas and their relations are summarized in Figure \ref{fig:overview}, where we can see that there are several ideas (like edit-distances, refinement operators or kernels) that apply across all representation formalisms. The advantage of these general ideas is that they are universal and can be applied to any type of data. For example, the same exact formulation of a refinement operator-based distance function can be applied to graphs, frames or logical expressions. However, the price to pay is computational complexity, as refinement operator distances, or edit distances are computationally very expensive. Thus the work on numerical approximations to these functions.

Figure \ref{fig:overview} also makes explicit how the different basic ideas of similarity for non-structured data types have influenced the work on structured distance and similarity functions. For example, the basic idea of {\em edge counting}, originally proposed for defining distance functions between elements of a taxonomy, when instantiated for different structured data representations gives rise to graph edit distances (where ``edge counting'' transforms into ``edit operation counting''), or antiunification-based distances (where we see the subsumption graph as a taxonomy and directly apply edge counting), among others.

Additionally, by looking at all of these distance and similarity concepts side by side, we can also identify other types of relations between them. For example, as we saw in Section \ref{sec:graph-distance}, the Jaccard index and the idea of edge counting in taxonomies are very related ideas: both count the size of what do two objects have in common (the intersection in the case of sets, or their common ancestor in the case of taxonomies). Thus, when instantiating these two ideas in data structures like graphs, they correspond to MCS-type functions.


Moreover, the key difference between each of these similarity or distance functions is their {\em bias}. The {\em bias} of a machine learning method (often called the {\em inductive bias}) is defined as the mechanisms and preferences that are intrinsic to a particular algorithm and that given some training data make it select a specific given hypothesis or model of the data from all the other equally good models in the hypothesis space~\citep{mitchell1980need}. In the same way, distance and similarity functions encode their own biases. For example, hierarchical aggregation methods for tree-based representations consider that the information that is deeper in the tree is less important that that on the shallower levels of the tree. While this could suit some domains, it might not suit others. Thus, it is important to understand the types of similarities and differences each function considers, and which biases it introduces, since one or another might be better suited for a particular application domain 
(as could be expected, given the {\em no free lunch} theorem~\citep{wolpert1996lack}). 

This is not unique to distance functions for structured data, as the same is true for classic distance functions. Consider, for example, the case of the Euclidean distance and the Cosine similarity. While both are designed to compare real valued vectors, the Cosine similarity is ``blind'' to the magnitude of the vectors and only considers their relative orientation. In some application domains, this is convenient, as the magnitude of the vectors might be irrelevant, but in some others the magnitude might be relevant, and thus Euclidean distance will be more appropriate. Thus, in summary, it is important to understand what is it that a given similarity or distance function is exactly measuring, as this will introduce a bias, which will suit some tasks but not others.

To this extend, functions that allow for {\em fitting}, i.e., those that contain parameters that can be trained given some training data, are interesting, since they can, to some extend adapt their implicit bias to specific domains (although, as is well known in machine learning, bias cannot be completely eliminated, as even the choice of knowledge representation used introduces a certain bias). This is not specific to structure representations. For example, in feature vectors, when deciding between when to use, say cosine similarity or Euclidean distance, the key is wether the magnitude of the vectors is important in our application domain (which is ignored by cosine similarity, but considered by Euclidean distance, thus introducing a different type of bias).

%
%
%


\section{Conclusions and Open Research Questions}\label{sec:conclusions}

This paper has presented an overview of existing work on distance and similarity functions for structured data representations. This is an important line of work, as data in many real world applications, such as in biomedical domains, is inherently structured. Specifically, we have organized the existing work along three types of representation formalisms: graphs, logic and frames, and discussed the different ideas existing in the literature concerning distance and similarity.

Despite the large body of work in structure similarity assessment, there are still a number of open research questions that need addressing. Some of these include:

{\bf Scalability}: many of the most powerful distance functions (such as edit distances and refinement operator-based ones), have a prohibitive computational complexity when dealing with either graph-based representations or complex logic-based ones. Although efficient approximations exist for some cases, this is not true in general. An interesting research direction would be the potential to exploit recent ideas of graph embeddings using neural networks to learn approximations to some of these distance functions, or to directly define fitted distance functions given a training set that would be efficient to calculate once the embedding network has been trained. An important related idea is that of {\em graph networks}~\citep{battaglia2018relational}, a family of neural networks designed to handle relational and graph data that has emerged over the past decade or so. Integrating classic ideas of distance and similarity with modern machine learning techniques might allow to scale up and harness very large amounts of data is thus a very promising future research direction. 
Promising results on this direction were recently published by \cite{li2019graph}, as discussed in Section \ref{subsec:graphnn}.

{\bf Cross-representation functions}: another open problem is that of defining general distance and similarity functions. Most of the work on distance function definition reported in this paper comes from separate communities (such as graph matching, inductive logic programming, machine learning, case-based reasoning). As a consequence, many of the ideas have been reinvented in these different fields. A unified theory of distance or similarity assessment for structured representations that could unify all of this work does not exist, and distance functions that are independent of the underlying representation formalism also do not exist.

{\bf Metric learning}: as discussed in several parts of this paper, different distance functions just capture different biases on assessing what is or not similar between two instances. However, without specifying a particular task at hand, choosing between one distance function or another is arbitrary, as different functions would be better suited for different tasks. Thus, research (such as metric learning) in defining distance functions that can be fitted to a given specific domain given training data is a very important research direction. Although work on metric learning for structured representations has started, more work is needed in order to have practical alternatives that can scale to large structured representations. Again, the idea of graph networks mentioned above can play an important role in future work in this direction.

{\bf Interpretability}: finally, many distance functions are black boxes, and it is hard to understand why have they produced a given distance value. While some work (e.g., that of \cite{plaza2005explanatory}) has worked on producing symbolic similarity values that are human interpretable, in general, most distance functions are still opaque. Research into how to explain predictions mace by distance function-based machine learning algorithms is thus needed.

\bibliographystyle{spbasic}

\end{document}